\documentclass[10pt,twocolumn,letterpaper]{article}
\usepackage{cvpr}      % To produce the REVIEW version
\usepackage[accsupp]{axessibility}
\definecolor{cvprblue}{rgb}{0.21,0.49,0.74}
\usepackage[breaklinks,colorlinks,allcolors=cvprblue]{hyperref}
\usepackage[ruled,linesnumbered]{algorithm2e} % algorithms
\usepackage{colortbl}      % colored table cells (use colortbl, NOT xcolor[table])
\usepackage{multirow}      % multi-row table cells
\usepackage{bm}            % bold math symbols
\usepackage{siunitx}       % number and unit formatting
\usepackage{pifont}        % \ding symbols (e.g. check / cross marks)
\usepackage{makecell}      % line breaks inside table cells
\usepackage{tabularx}      % auto-width tabulars
\usepackage{array}         % advanced column specs
\makeatletter

\newcommand{\Rmnum}[1]{\expandafter\@slowromancap\romannumeral #1@}
\makeatother

\definecolor{lightred}{rgb}{1.0, 0.78, 0.81}
\definecolor{lightpurple}{rgb}{0.88, 0.77, 1}

\definecolor{detgreen}{HTML}{00D030}
\definecolor{detpurple}{HTML}{902090}
\definecolor{detblue}{HTML}{1876E5}
\DeclareRobustCommand{\detbox}[1]{%
    \textcolor{#1}{\setlength{\unitlength}{1pt}%
    \begin{picture}(11,7)\linethickness{1.0pt}%
        \put(0,0.5){\dashbox{1.2}(10,5){}}%
    \end{picture}}%
}
\newlength{\TableMethodGap}
\newlength{\TablePublicationGap}
\newlength{\TableMetricGap}
\newlength{\TableDatasetGap}

\def\confName{CVPR}
\def\confYear{2027}
\title{GenRF: Generative Registration-free Image Fusion}
\author{Linfeng Tang\textsuperscript{1} \quad Zhijun Fan\textsuperscript{1} \quad Yang Chen\textsuperscript{2} \quad Zizhuo Li\textsuperscript{1} \quad Hao Zhang\textsuperscript{1} \quad Jiayi Ma\textsuperscript{1}\thanks{Corresponding author.}\\ \textsuperscript{1}Wuhan University, China \quad \textsuperscript{2}Raytron Technology, China}

\usepackage[breakable,skins]{tcolorbox}
\tcbset{fusionprompt/.style={enhanced,boxrule=0.4pt,arc=1.2mm,
 left=7pt,right=7pt,top=6pt,bottom=6pt,boxsep=0pt,
 before skip=7pt,after skip=8pt,fonttitle=\small\bfseries,
 fontupper=\small,coltitle=black,toptitle=4pt,bottomtitle=4pt}}
\newcommand{\clip}{\operatorname{clip}}
\definecolor{degvisible}{RGB}{56,125,176}
\definecolor{degshared}{RGB}{125,92,161}
\definecolor{deginfrared}{RGB}{66,145,107}
\definecolor{tablemodalityblue}{RGB}{128,128,255}
\usepackage{soul}
\colorlet{promptbackground}{degshared!5}
\newcommand{\inlineprompt}[1]{{\sethlcolor{promptbackground}\hl{#1}}}

\begin{document}

\maketitle
\begin{abstract}
Multi-modal image fusion combines complementary observations to improve scene visibility and perception. However, in real-world acquisition, differences in sensing principles and configurations introduce geometric discrepancies, while degradations further compromise source quality. Existing methods typically rely on explicit registration, yet large parallax complicates correspondence estimation, while mismatched fields of view leave some regions without cross-modal counterparts. Meanwhile, independent restoration underuses cross-modal information, whereas restoration–fusion methods assuming alignment remain vulnerable to misregistration. This study revisits misaligned and degraded image fusion in the era of generative foundation models, in which cross-attention enables position-agnostic complementary information aggregation, while generative priors support degradation recovery and inference in missing regions. Building on this perspective, we propose GenRF, which adapts pretrained FLUX through LoRA to jointly enhance degraded observations, infer missing cross-modal context, and fuse complementary information without explicitly estimating deformation fields. Extensive experiments on real-world infrared and visible captures show that GenRF delivers high-quality fusion directly from unregistered, degraded images without preprocessing, while improving downstream task performance under misalignment. Additional experiments demonstrate that GenRF can generalize to misaligned multi-exposure and multi-focus image fusion without task-specific fine-tuning. %Project: \url{https://linfeng-tang.github.io/GenRF}.
\end{abstract}
\section{Introduction}
\label{sec:intro}

Multi-modal image fusion is an important image enhancement technique that integrates complementary observations from heterogeneous sensors to provide a comprehensive description of the scene~\cite{Tang2023Survey, Liu2025Survey}. As a representative task, infrared and visible image fusion combines visible textures with infrared thermal information, which can improve target visibility under challenging imaging conditions. Such fused representations not only facilitate human visual observation~\cite{Tang2023DIVFusion}, but also enhance machine perception~\cite{Tang2022SeAFusion}.

\begin{figure}[t]
  \centering
  \includegraphics[width=0.478\textwidth]{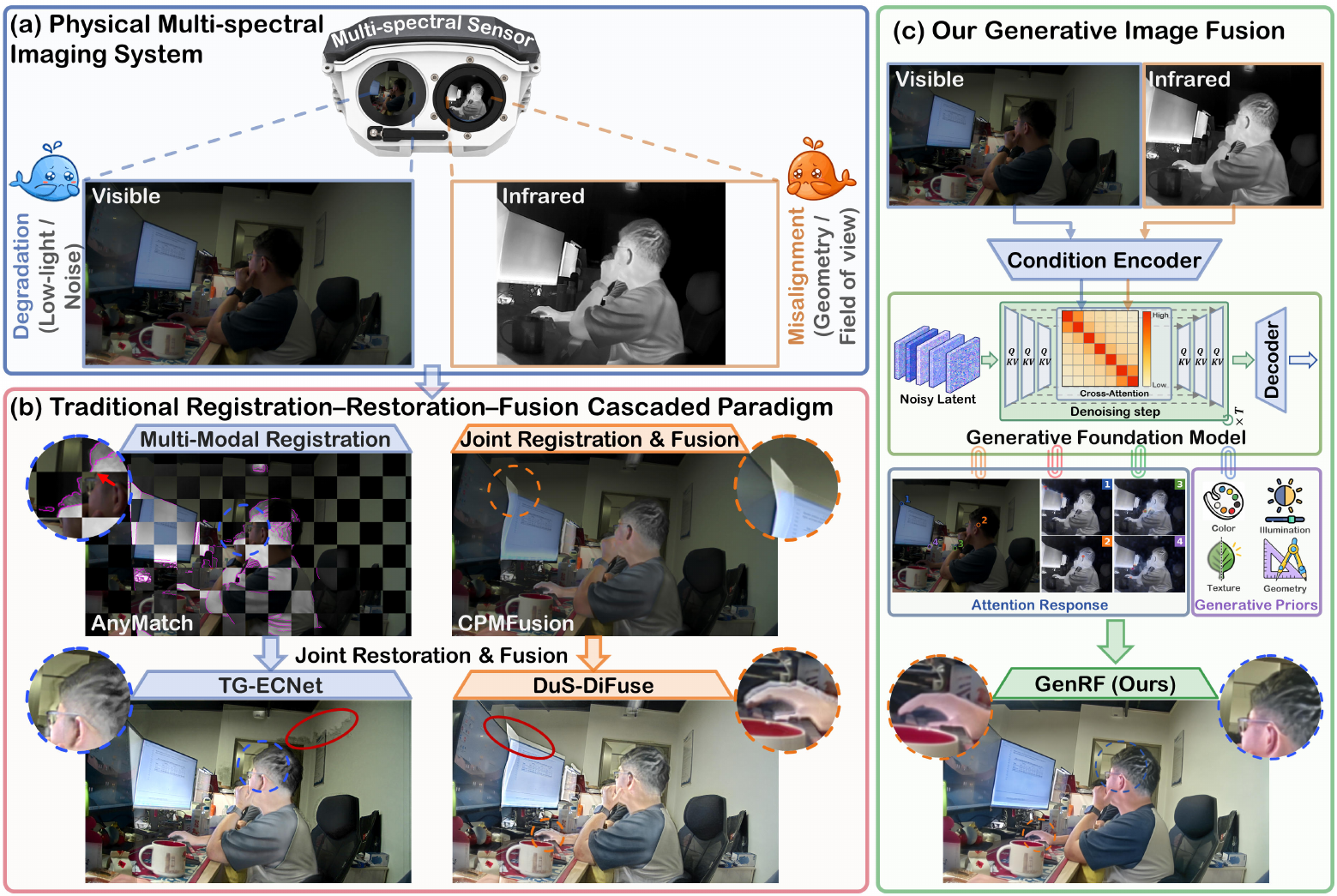}
  \caption{Real-world multi-modal acquisition challenges and representative solutions. TG-ECNet and DuS-DiFuse results are obtained using image pairs registered by AnyMatch and CPMFusion.}
  \vspace{-0.15in}
  \label{fig:example}
\end{figure}

However, the practical acquisition of multi-modal images usually introduces challenges beyond information aggregation. As illustrated in Fig.~\ref{fig:example}~(a), heterogeneous imaging characteristics and configurations result in geometric misalignment and partial field-of-view overlap~\cite{Yi2026SeeRF}. Meanwhile, low-light, noise, and low contrast further compromise source quality~\cite{Tang2023DIVFusion, Yi2024Text-IF, Zhang2025OmniFuse}. Degradation obscures structural cues and hinders cross-modal correspondence estimation, while geometric inconsistencies cause spatially mismatched content to be integrated, thereby introducing artifacts~\cite{Tang2025C2RF, Zhang2026G2Fusion}.

To address misalignment, existing solutions mainly follow two strategies, namely registration before fusion and joint registration--fusion optimization. The former establishes cross-modal correspondences with modality-agnostic descriptors, such as RIFT~\cite{Li2020RIFT} and ReDFeat~\cite{Deng2023ReDFeat}, or adapts matching models through multi-modal data generation, as in MINIMA~\cite{Ren2025MINIMA} and AnyMatch~\cite{Yang2026AnyMatch}. These correspondences are often utilized to estimate a homography for pre-registration. However, \emph{this approximation becomes inadequate when scene depth is insufficient relative to the camera baseline, as depth-dependent parallax cannot generally be represented by a single global transformation}~\cite{Zaragoza2013APAP}. Consequently, the background may be aligned while nearby targets remain displaced, introducing local artifacts during fusion, as illustrated in Fig.~\ref{fig:example}~(b). Therefore, joint registration and fusion methods introduce more flexible geometric modeling, progressively refining coarse alignment with local displacement estimation~\cite{Xu2023MURF,Liu2026URMIF} or directly predicting dense deformation fields~\cite{Tang2022SuperFusion,Tang2025C2RF,Xiao2026CPMFusion}. However, cross-modal appearance discrepancies and weak local structure leave dense correspondence estimation underconstrained. As shown in Fig.~\ref{fig:example}~(b), CPMFusion alleviates artifacts around salient targets but leaves visible distortions in smoother regions, such as the monitor. Beyond geometric misalignment, \emph{infrared and visible sensors often differ in field of view, leaving some regions without cross-modal counterparts}~\cite{Yi2026SeeRF}. Existing methods typically accommodate mismatched fields of view via constant-value padding, yet this operation cannot recover missing cross-modal information and may introduce pronounced artifacts.

Additionally, to address image degradation, the community has developed a series of joint restoration and fusion methods that exploit cross-modal complementarity to restore degraded content while integrating useful source information. Early efforts primarily targeted specific degradations, such as poor illumination~\cite{Tang2023DIVFusion} and noise interference~\cite{Liu2023PAIF}. More recent all-in-one methods handle diverse degradations in a unified model via degradation-aware feature modulation~\cite{Yi2024Text-IF,Tang2025ControlFusion}, expert routing~\cite{Sun2025TGECNet}, or generative modeling~\cite{Zhang2025OmniFuse, Cao2026DuSDiFuse, Shi2026DRFusion}. Notably, \emph{these methods generally assume aligned inputs, whereas real-world captures often exhibit geometric discrepancies}. Therefore, their practical deployment requires preceding registration, forming a cascaded registration--restoration--fusion pipeline in which residual alignment errors lead to distortions and artifacts. As shown in Fig.~\ref{fig:example}~(b), applying DuS-DiFuse to CPMFusion-registered results amplifies residual misaligned content, causing ghosting to re-emerge around the head. 

The aforementioned observations invite us to rethink: \emph{does complementary information aggregation necessarily require explicit registration?} Generative foundation models offer a promising alternative via two key components, \emph{i.e.,} cross-attention and generative priors. On the one hand, attention supports content-dependent retrieval across spatial positions~\cite{Chen20263MTI}, allowing complementary features to be aggregated without requiring them to occupy the same pixel, as presented in Fig.~\ref{fig:example}~(c). On the other hand, large-scale training on natural images equips generative models with appearance and structural priors that promote balanced exposure, natural colors, and fine textures in fusion results~\cite{Cao2026DuSDiFuse}. In particular, structural priors help infer content outside one sensor's field of view, while texture priors may help suppress artifacts and thereby guide cross-attention toward coherent information aggregation during fine-tuning.

Building on this perspective, we propose \textbf{GenRF}, a generative registration-free image fusion framework that fine-tunes pretrained FLUX~\cite{BFL2026Flux3} via low-rank adaptation (LoRA)~\cite{Hu2022LoRA}. To unlock the potential of cross-attention under misalignment, we develop an online geometric perturbation strategy that combines global transformations with local elastic deformations. Meanwhile, we combine real degraded observations with dynamically synthesized corruptions to harness pretrained generative priors for restoration under diverse degradations. Finally, misaligned observations serve as joint conditions for iterative denoising, yielding enhanced fusion results in the visible reference geometry without explicitly estimating deformation fields or applying a separate restoration step. As illustrated in Fig.~\ref{fig:example}~(c), GenRF produces a visually coherent fusion result free of degradation and ghosting. Notably, \textit{there are no visible artifacts even in regions outside the overlapping field of view}. In summary, our main contributions are as follows:
\begin{itemize}
  \item We revisit misaligned and degraded multi-modal image fusion under a generative paradigm, where cross-attention retrieves complementary information beyond pixelwise correspondence and generative priors support degradation recovery and missing-content inference.
  \item We propose GenRF, a generative registration-free framework that adapts pretrained FLUX via LoRA to directly fuse real-world captures without separate registration or restoration. Geometric perturbations help cross-attention adapt to misalignment, while degradation modeling harnesses pretrained generative priors for restoration.
  \item Extensive experiments on real captures and four public benchmarks demonstrate superior fusion quality, geometric consistency, and degradation suppression, with benefits for downstream perception. \textit{Extended results on misaligned multi-exposure and multi-focus fusion without fine-tuning confirm the generalizability of GenRF}.
\end{itemize}
\section{Related Work}
\subsection{Multi-modal Image Registration and Fusion}
Multi-modal image registration seeks reliable correspondences under significant cross-modal appearance variations. Descriptor-based approaches address this challenge via hand-crafted or learned modality-agnostic representations, such as RIFT~\cite{Li2020RIFT} and ReDFeat~\cite{Deng2023ReDFeat}. Recently, data-driven approaches, \emph{e.g.}, MINIMA~\cite{Ren2025MINIMA} and AnyMatch~\cite{Yang2026AnyMatch}, adapt matching networks to heterogeneous observations~via multi-modal data generation. Resulting correspondences support geometric transformation estimation and provide pre-registration for subsequent fusion, yet correspondence learning remains separate from information integration.

Joint registration and fusion methods establish cross-task interactions to exploit fusion feedback for alignment refinement. Early efforts reduce modality discrepancies with image translation~\cite{Wang2022UMF} or shared representations~\cite{Xu2023MURF}, followed by coarse-to-fine alignment and fusion. Furthermore, SuperFusion~\cite{Tang2022SuperFusion} strengthens fusion feedback via consistency constraints on symmetric fusion outputs. Additionally, C2RF~\cite{Tang2025C2RF} introduces contrastive learning between aligned and misaligned fusion results, thus making registration optimization more sensitive to misalignment artifacts.

More recent work advances joint registration and fusion through unified architectures and geometric modeling. In particular, AU-Net~\cite{Lu2025AUNet} integrates the two tasks within a shared network, reducing the separation between their feature processing stages. In terms of geometric modeling, CPMFusion~\cite{Xiao2026CPMFusion} exploits modal correlations for bidirectional registration, whereas URMIF~\cite{Liu2026URMIF} combines structural representations with a dominant-plane prior. 

However, mismatched fields of view introduce a further challenge, as some scene regions are observed by only one sensor. SeeRF~\cite{Yi2026SeeRF} is the first work to address this limitation through flow-guided registration and diffusion-based completion, using a validity mask to identify non-overlapping regions. Notably, a global homography cannot capture depth-dependent parallax~\cite{Zaragoza2013APAP}, while reliable cross-modal correspondences are often sparse, leaving dense flow estimation insufficiently constrained. In contrast, GenRF uses cross-attention for complementary information aggregation without explicit registration, while exploiting generative priors to infer missing content.

\subsection{Joint Image Restoration and Fusion}
Joint restoration and fusion methods exploit cross-modal complementarity to recover degraded content while integrating useful information. Early studies improved the robustness of fusion models through degradation-specific designs. DIVFusion~\cite{Tang2023DIVFusion} disentangles illumination from scene content to mitigate low-light degradation, while PAIF~\cite{Liu2023PAIF} combines architecture search with adaptive adversarial learning to improve robustness to noise interference. Deno-IF~\cite{Xu2025DenoIF} further employs convolutional low-rank priors for joint denoising and fusion.

To accommodate diverse degradations, all-in-one methods introduce degradation-aware modeling and adaptive feature processing within a unified framework. DAFusion~\cite{Wang2025DAFusion} learns degradation-aware representations through unpaired contrastive supervision, whereas URFusion~\cite{Xu2025URFusion} separates degradation-free intrinsic content from appearance to enable unsupervised restoration and fusion. For more explicit degradation guidance, Text-IF~\cite{Yi2024Text-IF} and ControlFusion~\cite{Tang2025ControlFusion} use text or language-vision prompts to modulate feature processing. To eliminate reliance on textual prompts, DSPFusion~\cite{Tang2026DSPFusion} combines contrastively learned degradation priors with diffusion-restored semantic priors to guide restoration and fusion. In addition, TG-ECNet~\cite{Sun2025TGECNet} coordinates restoration and fusion through expert routing.

To recover high-quality content from degraded observations, generative approaches leverage learned image priors to guide restoration and fusion. In particular, DRMF~\cite{Tang2024DRMF} composes modality-specific diffusion priors for complementary information integration. OmniFuse~\cite{Zhang2025OmniFuse} extends this direction to composite degradations with latent diffusion modeling, while DuS-DiFuse~\cite{Cao2026DuSDiFuse} incorporates controllable generative modulation for restoration and enhancement. Moreover, ReCoFuse~\cite{Zhang2026ReCoFuse} further couples modality-specific restoration with cross-modal integration through reciprocal feedback. Notably, despite improved degradation robustness, these methods assume aligned observations.

To handle misalignment and degradation together, G2Fusion~\cite{Zhang2026G2Fusion} combines flow-based registration with DiT-based generative fusion in a geometric-to-generative framework. Fusion feedback further refines registration, enabling coordinated optimization of the three tasks rather than independent cascaded processing. Nevertheless, this framework still relies on explicit deformation estimation and shows limited generalization to real captures, where residual misalignment leads to pronounced artifacts in the fusion results. In contrast, GenRF directly conditions generation on misaligned observations, integrating restoration and fusion without a separate registration stage.
\begin{figure*}[t]
  \centering
  \includegraphics[width=0.95\textwidth]{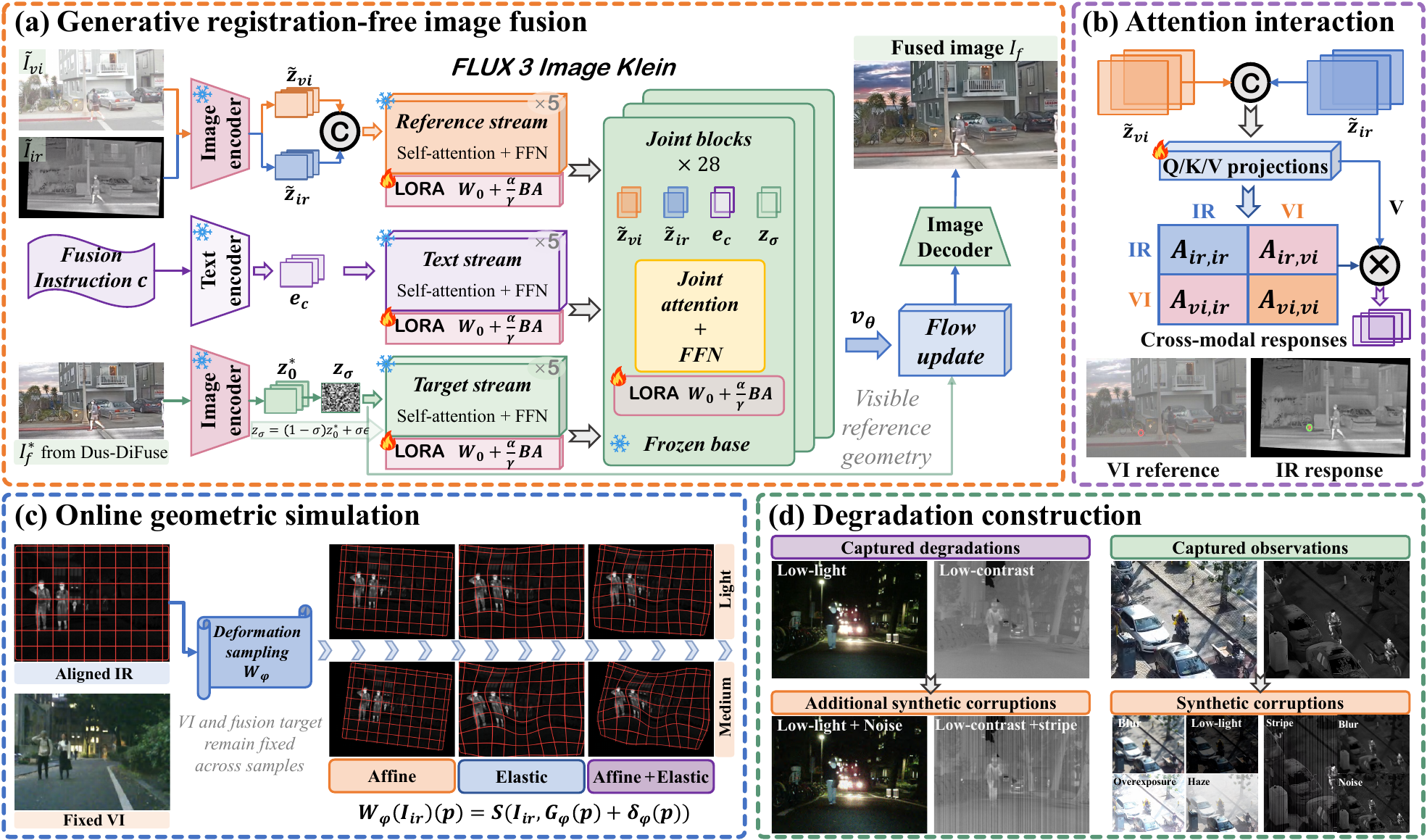}
  \vspace{-0.05in}
  \caption{Overall framework of our generative registration-free image fusion.}
  \vspace{-0.08in}
  \label{fig:framework}
\end{figure*}

\section{Methodology}
\label{sec:methodology}
\subsection{Generative Formulation}
Given misaligned infrared and visible observations $I_{ir}$ and $I_{vi}$, we aim to generate a high-quality fused image $I_f$ in the visible reference geometry. The observations may suffer from various degradations, \emph{e.g.}, low-light, noise, blur, and low-contrast, while covering only partially overlapping fields of view. Instead of first estimating a geometric transformation, we formulate misaligned image fusion as conditional generation, $I_f \sim p_{\theta}(I_f\mid I_{ir},I_{vi},c)$, where $c$ is a fusion instruction. In the generation process, cross-attention aggregates complementary context across misaligned spatial positions, while generative priors support restoration and content inference when observations are incomplete.

As shown in Fig.~\ref{fig:framework}~(a), we instantiate this formulation with pretrained FLUX 3~\cite{BFL2026Flux3}. First, the frozen VAE encoder $\mathcal E$ and text encoder $\mathcal E_c$ encode source images and instruction:
\begin{equation}
    z_{ir}=\mathcal E(I_{ir}),\quad
    z_{vi}=\mathcal E(I_{vi}),\quad
    e_c=\mathcal E_c(c).
    \label{eq:genrf-conditioning}
\end{equation}

Generation starts from Gaussian noise $z_1\sim\mathcal N(0,I)$. We denote the evolving target latent at noise level $\sigma\in[0,1]$ by $z_\sigma$, with $z_0$ representing the fused-image latent. At each noise level, the transformer predicts the conditional velocity of $z_\sigma$ from the source latents and text embedding. Generation integrates these velocity predictions to transform $z_1$ into the fused-image latent $z_0$,
\begin{equation}
    \frac{\mathrm d z_\sigma}{\mathrm d\sigma}
    =v_\theta(z_\sigma,\sigma;z_{ir},z_{vi},e_c),
    \quad I_f=\mathcal D(z_0),
    \label{eq:genrf-generation}
\end{equation}
\(\sigma\) decreases from \(1\) to \(0\), and \(\mathcal D\) is the frozen VAE decoder.

In order to adapt the pretrained generative priors for multi-modal image fusion, we apply LoRA~\cite{Hu2022LoRA} to the attention projections and feed-forward layers:
\begin{equation}
    W=W_0+\frac{\alpha}{r}BA,
    \label{eq:genrf-lora}
\end{equation}
where $W_0$ denotes the frozen weight matrix, $A$ and $B$ are trainable low-rank matrices, $r$ determines the low-rank dimension, and $\alpha$ controls the update scale. Fusion supervision updates only the low-rank matrices for task adaptation. Thus, the frozen backbone retains generative priors learned from large-scale natural images, such as balanced illumination, natural colors, abundant textures and structures.

\subsection{Attention-based Information Aggregation}
\begin{figure}[t]
    \centering
    \includegraphics[width=0.99\linewidth]{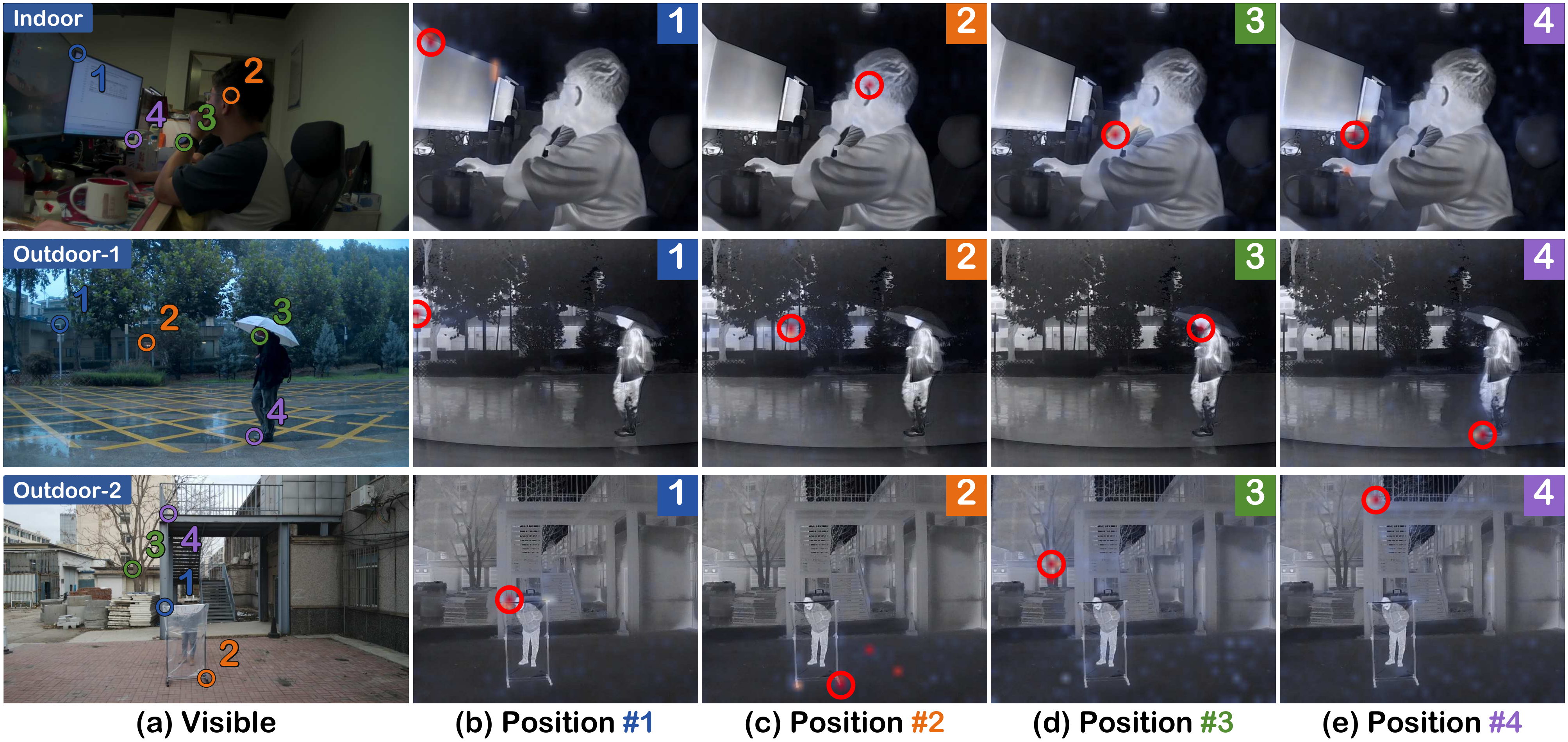}
    \vspace{-0.1in}
	\caption{Attention responses in infrared images for numbered visible queries. \textcolor{red}{\(\bm{\circ}\)} marks each averaged attention map's peak.}
    \vspace{-0.1in}
    \label{fig:attention-response}
\end{figure}

FLUX 3 adopts a two-stage transformer architecture with stream-specific processing followed by joint interaction. The first stage processes image conditions, text, and the noisy target $z_\sigma$ in three separate streams. The infrared and visible latents are tokenized and concatenated along the sequence dimension to form the shared image-conditioning stream. Self-attention over this sequence supports both within-modal and cross-modal interactions~\cite{Chen20263MTI}, as shown in Fig.~\ref{fig:framework}~(b). The unified output of a single attention head is:
\begin{equation}
    O_r=
    \underbrace{\begin{bmatrix}
        \mathcal A_{ir,ir}&\mathcal A_{ir,vi}\\
        \mathcal A_{vi,ir}&\mathcal A_{vi,vi}
    \end{bmatrix}}_{\mathcal A_r}
    \begin{bmatrix}V_{ir}\\V_{vi}\end{bmatrix},
    \label{eq:genrf-reference-attention}
\end{equation}
where $\mathcal A_r=\operatorname{softmax}(Q_rK_r^\top/\sqrt d)$ is the joint attention matrix and $d$ is the head dimension. The queries $Q_r$, keys $K_r$, and values $V_r$ are derived from the concatenated sequence, with $V_r=[V_{ir};V_{vi}]$. The diagonal blocks $\mathcal A_{ir,ir}$ and $\mathcal A_{vi,vi}$ model within-modal self-attention, while $\mathcal A_{ir,vi}$ and $\mathcal A_{vi,ir}$ enable bidirectional cross-modal attention:
\begin{equation}
    \mathcal A_{m,n}(i,j)=
    \frac{\exp\!\left(q_{m,i}^{\top}k_{n,j}/\sqrt d\right)}
    {\displaystyle\sum_{u\in\{ir,vi\}}\sum_{t=1}^{N_u}
    \exp\!\left(q_{m,i}^{\top}k_{u,t}/\sqrt d\right)},
    \label{eq:genrf-cross-modal-weights}
\end{equation}
where $m,n\in\{ir,vi\}$ index the query/key modalities, $q_{m,i}$ and $k_{n,j}$ are the query/key vectors used in attention, and $N_u$ counts tokens in modality $u$. In particular, $\mathcal A_{vi,ir}(i,j)$ allows visible token $i$ to query any infrared token $j$ without a prescribed geometric mapping. Thus, complementary information can be aggregated through learned affinities across misaligned spatial positions. As shown in Fig.~\ref{fig:attention-response}, the queries in visible images produce localized attention peaks in corresponding infrared regions despite spatial misalignment. Suppl. Material compares attention peaks with known correspondences under simulated deformations.

In the second stage, the image-conditioning, text, and noisy-target streams are concatenated and processed by joint transformer blocks. The image conditions provide scene context and thermal cues, while text features specify the desired fusion characteristics. Through joint attention, the evolving target queries complementary source information under the fusion instruction. %The resulting target features are projected to predict the velocity field in Eq.~\eqref{eq:genrf-generation}.

\begin{figure*}[!t]
    \centering
    \includegraphics[width=0.99\textwidth]{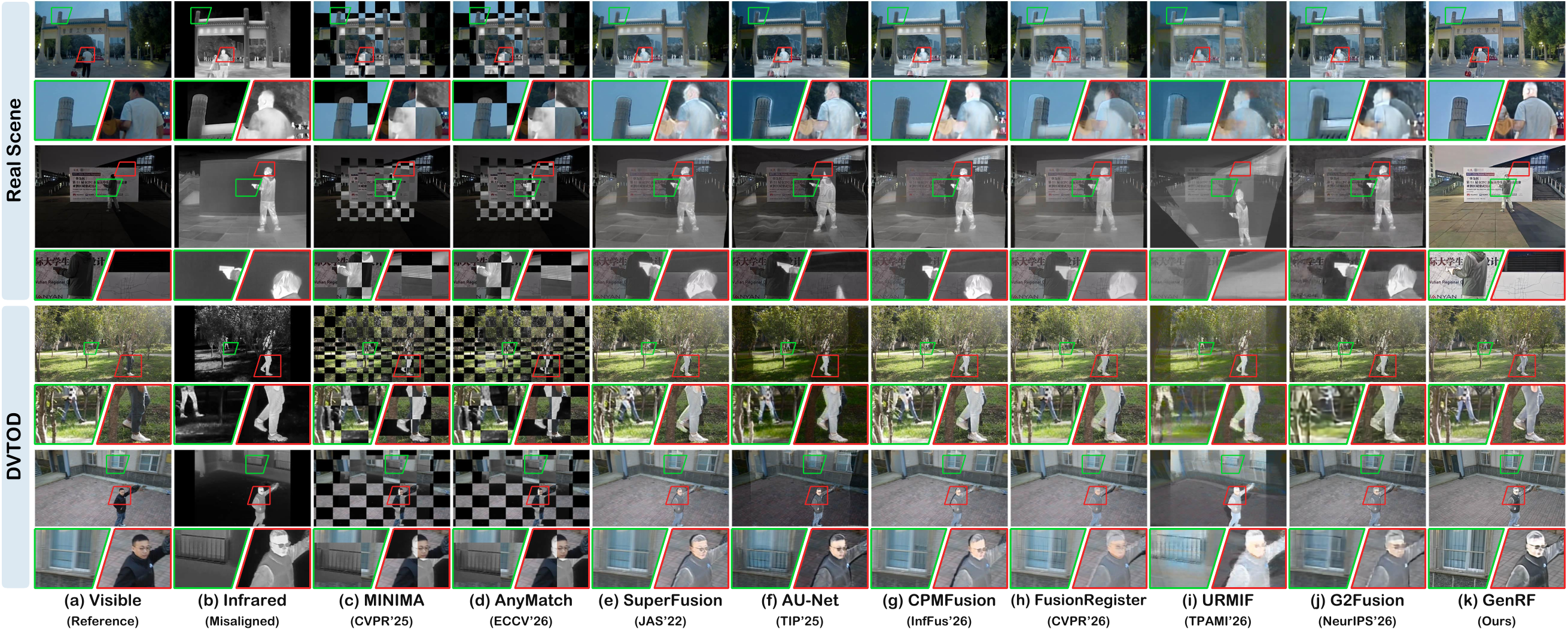}
    \vspace{-0.06in}
    \caption{Qualitative comparison with registration-based methods on Real Scene and DVTOD. The first Real Scene example was captured with a rigidly mounted dual-sensor camera with a narrow baseline yet exhibits large scene-depth variation. The second uses a separate infrared camera and phone with a larger field-of-view mismatch. The first DVTOD example additionally exhibits temporal misalignment.}
    \label{fig:registration-comparison}
\end{figure*}

\subsection{Learning from Perturbed Observations}
To enhance the ability of cross-attention for aggregating complementary information under misalignment and adapt generative priors for degradation recovery, we introduce an online geometric perturbation and dynamic degradation construction strategy. In particular, given the aligned high-quality source images $I_{ir}^{\mathrm{ref}}$ and $I_{vi}^{\mathrm{ref}}$, DuS-DiFuse~\cite{Cao2026DuSDiFuse} is employed to generate the desired fused image $I_f^\star$. We then perturb the source conditions while keeping $I_f^\star$ fixed.

\textbf{Geometric perturbations.} Training exclusively on aligned pairs biases cross-attention toward aggregating complementary information at matching coordinates, limiting its flexibility under misalignment. Therefore, we construct perturbed samples from aligned $I_{ir}^{\mathrm{ref}}$ and $I_{vi}^{\mathrm{ref}}$ by applying a sampled warp $\mathcal W_\phi$ to the infrared image. As shown in Fig.~\ref{fig:framework}~(c), the warp combines global transformations with local elastic displacements~\cite{Tang2022SuperFusion,Tang2025C2RF}, and is formulated as:
\begin{equation}
    \mathcal W_\phi(I_{ir}^{\mathrm{ref}})(\mathbf p)
    =\mathcal S\!\left(I_{ir}^{\mathrm{ref}},g_\phi(\mathbf p)+\delta_\phi(\mathbf p)\right),
    \label{eq:genrf-geometric-warp}
\end{equation}
where $\mathbf p$ is an output-grid location and $\mathcal S$ denotes the resampling operation. $g_\phi$ and $\delta_\phi$ denote the global transformation and local elastic displacement. Both are sampled online to simulate diverse misalignments for better generalization to real captures. The warp also moves some infrared content outside the image domain, leaving uncovered regions that naturally simulate field-of-view mismatch.

\textbf{Degradation construction.} To improve robustness to diverse degradations in real captures, we develop an online degradation generator that varies degradation types and strengths, as shown in Fig.~\ref{fig:framework}~(d). We utilize the high-quality images from DSPFusion~\cite{Tang2026DSPFusion} as references and extend its degradation simulation with additional degradation types and online parameter sampling. Let $\mathcal D_m$ denote degradation construction for modality $m$. The source conditions are:
\begin{equation}
    \widetilde I_{ir}=\mathcal W_\phi\!\left(\mathcal D_{ir}(I_{ir}^{\mathrm{ref}};\eta_{ir})\right),\quad
    \widetilde I_{vi}=\mathcal D_{vi}(I_{vi}^{\mathrm{ref}};\eta_{vi}),
    \label{eq:genrf-conditions}
\end{equation}
where $\eta_m$ specifies degradation type and strength. $\mathcal D_{vi}$ incorporates low-light, over-exposure, motion blur, and weather interference, while $\mathcal D_{ir}$ includes low-contrast and stripes. Both modalities undergo noise and Gaussian blur. Alongside synthetic corruptions, we incorporate real low-light and over-exposed visible images and low-contrast infrared images to improve adaptation to real captures. More details are provided in Suppl. Material.

\begin{table*}[t]
	\centering
	\caption{Quantitative comparison on Real Scene and DVTOD. The best and second-best results are highlighted in \textcolor[HTML]{B95161}{red} and \textcolor[HTML]{6E6EA1}{purple}. }
	\vspace{-0.08in}
	\label{tab:real-scene-dvtod}
	\setlength{\TableMethodGap}{6pt}
	\setlength{\TablePublicationGap}{12pt}
	\setlength{\TableMetricGap}{12pt}
	\setlength{\TableDatasetGap}{15pt}
	\setlength{\tabcolsep}{5pt}
	\renewcommand{\arraystretch}{0.95}
	\small
	\resizebox{0.99\textwidth}{!}{%
		\begin{tabular}{@{}l@{}p{\TableMethodGap}@{}c@{}p{\TablePublicationGap}@{}c@{}p{\TableMetricGap}@{}c@{}p{\TableMetricGap}@{}c@{}p{\TableMetricGap}@{}c@{}p{\TableDatasetGap}@{}c@{}p{\TableMetricGap}@{}c@{}p{\TableMetricGap}@{}c@{}p{\TableMetricGap}@{}c@{}}
			\toprule
			\multirow{2}{*}{\textbf{Methods}} & & \multirow{2}{*}{\textbf{Publication}} & & \multicolumn{7}{c}{\textbf{Real Scene}} & & \multicolumn{7}{c}{\textbf{DVTOD}} \\
			\cmidrule(lr){5-11}\cmidrule(l){13-19}
			& &  & & EN~$\uparrow$ & & DeQA~$\uparrow$ & & LIQE~$\uparrow$ & & QualiCLIP~$\uparrow$ & & EN~$\uparrow$ & & DeQA~$\uparrow$ & & LIQE~$\uparrow$ & & QualiCLIP~$\uparrow$ \\
			\midrule
			\textbf{SuperFusion}~\cite{Tang2022SuperFusion} &  & \emph{JAS'22} &  & 7.119 &  & 2.764 &  & 2.222 &  & 0.541 &  & 7.232 &  & 3.080 &  & 2.589 &  & 0.525 \\
			\textbf{AU-Net}~\cite{Lu2025AUNet} &  & \emph{TIP'25} &  & 7.018 &  & 2.742 &  & 2.091 &  & 0.506 &  & 6.979 &  & 2.825 &  & 2.266 &  & 0.471 \\
			\textbf{MulFS-CAP}~\cite{Li2025MulFSCAP} &  & \emph{TPAMI'25} &  & 6.416 &  & 2.518 &  & 1.592 &  & 0.334 &  & 6.576 &  & 2.680 &  & 2.162 &  & 0.356 \\
			\textbf{CPMFusion}~\cite{Xiao2026CPMFusion} &  & \emph{InfFus'26} &  & 7.171 &  & 2.745 &  & 2.028 &  & 0.519 &  & 7.286 &  & 3.039 &  & 2.434 &  & 0.539 \\
			\textbf{FusionRegister}~\cite{Bian2026FusionRegister} &  & \emph{CVPR'26} &  & 7.108 &  & 2.637 &  & 1.703 &  & 0.435 &  & 7.228 &  & 2.923 &  & 2.178 &  & 0.547 \\
			\textbf{URMIF}~\cite{Liu2026URMIF} &  & \emph{TPAMI'26} &  & 6.802 &  & 2.042 &  & 1.031 &  & 0.265 &  & 6.843 &  & 2.054 &  & 1.082 &  & 0.257 \\
			\midrule
			\textbf{Text-IF}~\cite{Yi2024Text-IF} &  & \emph{CVPR'24} &  & 7.122 &  & 2.798 &  & 2.193 &  & 0.562 &  & 7.368 &  & 3.095 &  & 2.762 &  & 0.591 \\
			\textbf{TG-ECNet}~\cite{Sun2025TGECNet} &  & \emph{ICML'25} &  & \cellcolor[HTML]{E6E6FA}7.279 &  & 2.737 &  & 2.085 &  & 0.454 &  & \cellcolor[HTML]{E6E6FA}7.397 &  & 2.886 &  & 2.628 &  & 0.448 \\
			\textbf{PDFuse}~\cite{Cao2025PDFuse} &  & \emph{NeurIPS'25} &  & 7.208 &  & 2.706 &  & 1.806 &  & 0.414 &  & 7.375 &  & 2.828 &  & 2.121 &  & 0.417 \\
			\textbf{OmniFuse}~\cite{Zhang2025OmniFuse} &  & \emph{TPAMI'25} &  & 7.127 &  & 2.501 &  & 1.511 &  & 0.433 &  & 7.251 &  & 2.697 &  & 1.928 &  & 0.463 \\
			\textbf{DSPFusion}~\cite{Tang2026DSPFusion} &  & \emph{TIP'26} &  & 7.180 &  & 2.834 &  & 2.338 &  & 0.560 &  & \cellcolor[HTML]{FFC7CE}\textbf{7.430} &  & 2.867 &  & 2.540 &  & 0.573 \\
			\textbf{ReCoFuse}~\cite{Zhang2026ReCoFuse} &  & \emph{CVPR'26} &  & 7.266 &  & 2.569 &  & 1.613 &  & 0.477 &  & 7.344 &  & 2.829 &  & 2.192 &  & 0.527 \\
			\textbf{DiTFuse}~\cite{Li2026DiTFuse} &  & \emph{TPAMI'26} &  & 6.934 &  & \cellcolor[HTML]{E6E6FA}2.988 &  & \cellcolor[HTML]{E6E6FA}2.657 &  & 0.608 &  & 7.166 &  & \cellcolor[HTML]{E6E6FA}3.177 &  & 3.055 &  & 0.603 \\
			\textbf{DuS-DiFuse}~\cite{Cao2026DuSDiFuse} &  & \emph{TPAMI'26} &  & 7.213 &  & 2.898 &  & 2.631 &  & \cellcolor[HTML]{E6E6FA}0.664 &  & 7.350 &  & 3.070 &  & \cellcolor[HTML]{E6E6FA}3.088 &  & \cellcolor[HTML]{E6E6FA}0.680 \\
			\midrule
			\textbf{G2Fusion}~\cite{Zhang2026G2Fusion} &  & \emph{NeurIPS'26} &  & 7.100 &  & 2.421 &  & 1.578 &  & 0.353 &  & 7.218 &  & 2.627 &  & 1.853 &  & 0.382 \\
			\textbf{GenRF (Ours)} &  & -- &  & \cellcolor[HTML]{FFC7CE}\textbf{7.284} &  & \cellcolor[HTML]{FFC7CE}\textbf{3.213} &  & \cellcolor[HTML]{FFC7CE}\textbf{3.206} &  & \cellcolor[HTML]{FFC7CE}\textbf{0.708} &  & 7.273 &  & \cellcolor[HTML]{FFC7CE}\textbf{3.349} &  & \cellcolor[HTML]{FFC7CE}\textbf{3.229} &  & \cellcolor[HTML]{FFC7CE}\textbf{0.697} \\
			\bottomrule
		\end{tabular}%
	}
	\vspace{-0.1in}
\end{table*}

\subsection{Training Objective}
We fine-tune pretrained FLUX using conditional flow matching in latent space. Given the desired fused image $I_f^\star$, the frozen encoder $\mathcal E$ produces the target latent $z_0^\star=\mathcal E(I_f^\star)$. We sample $\epsilon\sim\mathcal N(0,I)$ and construct the noisy target $z_\sigma=(1-\sigma)z_0^\star+\sigma\epsilon$, where $\sigma\in[0,1]$ is the noise level. The derivative with respect to $\sigma$ gives the target velocity $\epsilon-z_0^\star$. Conditioned on the perturbed source latents $\widetilde z_m=\mathcal E(\widetilde I_m)$ for $m\in\{ir,vi\}$ and the fusion instruction embedding $e_c$, the model predicts this velocity by minimizing:
\begin{equation}
    \mathcal L_{\mathrm{FM}}=\mathbb E\!\left[\left\|v_\theta(z_\sigma,\sigma;\widetilde z_{ir},\widetilde z_{vi},e_c)-(\epsilon-z_0^\star)\right\|_2^2\right],
    \label{eq:genrf-fm}
\end{equation}
where the squared error is averaged over latent elements. 

For each scene, we supervise the model with the same target latent under different geometric and degradation perturbations. This encourages the model to aggregate complementary cues across misaligned positions and restore degraded content while preserving the visible reference geometry. At inference, misaligned real-world captures jointly condition the generation of a fused image in the same geometry, without explicit registration or separate restoration.
\section{Experiments}

\subsection{Implementation and Configuration Details}
We implement GenRF in PyTorch using pretrained FLUX 3 Image Klein. Training uses 2,071 image pairs from the dataset constructed by DSPFusion~\cite{Tang2026DSPFusion}, with the desired fused images $I_f^\star$ generated by DuS-DiFuse~\cite{Cao2026DuSDiFuse}. The LoRA rank and scaling factor are both set to 16. Optimization is performed for 16 epochs using AdamW with a learning rate of $1\times10^{-4}$ on four NVIDIA RTX 4090 GPUs, with a batch size of 2 per GPU. Inference uses 30 denoising steps with a classifier-free guidance scale of 3.5.

The compared methods include two multi-modal matching methods, \emph{i.e.,} MINIMA~\cite{Ren2025MINIMA} and AnyMatch~\cite{Yang2026AnyMatch}; six joint registration and fusion methods, \emph{i.e.,} SuperFusion~\cite{Tang2022SuperFusion}, AU-Net~\cite{Lu2025AUNet}, MulFS-CAP~\cite{Li2025MulFSCAP}, CPMFusion~\cite{Xiao2026CPMFusion}, FusionRegister~\cite{Bian2026FusionRegister}, and URMIF~\cite{Liu2026URMIF}; eight joint restoration and fusion methods, \emph{i.e.,} Text-IF~\cite{Yi2024Text-IF}, TG-ECNet~\cite{Sun2025TGECNet}, PDFuse~\cite{Cao2025PDFuse}, OmniFuse~\cite{Zhang2025OmniFuse}, DSPFusion~\cite{Tang2026DSPFusion}, ReCoFuse~\cite{Zhang2026ReCoFuse}, DiTFuse~\cite{Li2026DiTFuse}, and DuS-DiFuse~\cite{Cao2026DuSDiFuse}; and G2Fusion~\cite{Zhang2026G2Fusion} for joint registration, restoration, and fusion. We evaluate GenRF on our Real Scene collection of 250 image pairs from multiple dual-sensor imaging systems and four public datasets, \emph{i.e.,} DVTOD~\cite{Song2024DVTOD}, MSRS~\cite{Tang2022PIAFusion}, LLVIP~\cite{Jia2021LLVIP}, and M3FD~\cite{Liu2022TarDAL}. Fusion quality is assessed with EN, DeQA~\cite{You2025DeQA}, LIQE~\cite{Zhang2023LIQE}, and QualiCLIP~\cite{Agnolucci2024QualiCLIP}. 

\begin{figure*}[t]
    \centering
    \includegraphics[width=0.99\textwidth]{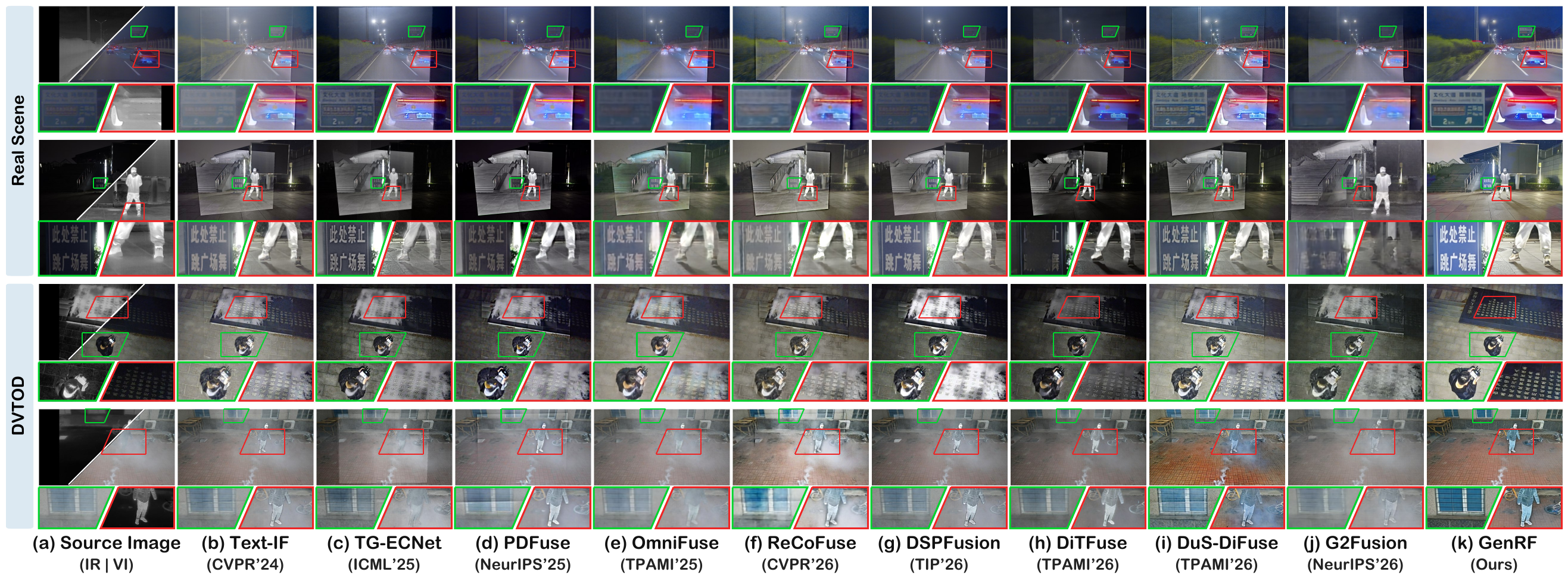}
    \vspace{-0.05in}
    \caption{Qualitative fusion on degraded Real Scene and DVTOD inputs. Restoration--fusion baselines use AnyMatch pre-registration.}
    \label{fig:real-fusion-comparison}
\end{figure*}

\begin{table*}[t]
	\centering
	\caption{Quantitative comparison with joint restoration and fusion alternatives on MSRS, LLVIP, and M3FD.}
	\label{tab:msrs-llvip-m3fd}
	\vspace{-0.05in}
	\setlength{\TableMethodGap}{4pt}
	\setlength{\TablePublicationGap}{10pt}
	\setlength{\TableMetricGap}{8pt}
	\setlength{\TableDatasetGap}{10pt}
	\setlength{\tabcolsep}{3pt}
	\renewcommand{\arraystretch}{1.}
	\small
	\resizebox{0.99\textwidth}{!}{%
		\begin{tabular}{@{}l@{}p{\TableMethodGap}@{}c@{}p{\TablePublicationGap}@{}c@{}p{\TableMetricGap}@{}c@{}p{\TableMetricGap}@{}c@{}p{\TableMetricGap}@{}c@{}p{\TableDatasetGap}@{}c@{}p{\TableMetricGap}@{}c@{}p{\TableMetricGap}@{}c@{}p{\TableMetricGap}@{}c@{}p{\TableDatasetGap}@{}c@{}p{\TableMetricGap}@{}c@{}p{\TableMetricGap}@{}c@{}p{\TableMetricGap}@{}c@{}}
			\toprule
			\multirow{2}{*}{\textbf{Methods}} & & \multirow{2}{*}{\textbf{Publication}} & & \multicolumn{7}{c}{\textbf{MSRS}} & & \multicolumn{7}{c}{\textbf{LLVIP}} & & \multicolumn{7}{c}{\textbf{M3FD}} \\
			\cmidrule(lr){5-11}\cmidrule(lr){13-19}\cmidrule(l){21-27}
			& &  & & EN~$\uparrow$ & & DeQA~$\uparrow$ & & LIQE~$\uparrow$ & & QualiCLIP~$\uparrow$ & & EN~$\uparrow$ & & DeQA~$\uparrow$ & & LIQE~$\uparrow$ & & QualiCLIP~$\uparrow$ & & EN~$\uparrow$ & & DeQA~$\uparrow$ & & LIQE~$\uparrow$ & & QualiCLIP~$\uparrow$ \\
			\midrule
			\textbf{Text-IF}~\cite{Yi2024Text-IF} &  & \emph{CVPR'24} &  & 7.016 &  & 2.229 &  & 1.387 &  & 0.360 &  & 7.362 &  & 2.454 &  & 2.077 &  & 0.629 &  & 6.835 &  & \cellcolor[HTML]{E6E6FA}2.660 &  & 2.099 &  & 0.522 \\
			\textbf{TG-ECNet}~\cite{Sun2025TGECNet} &  & \emph{ICML'25} &  & 6.325 &  & 2.083 &  & 1.319 &  & 0.266 &  & 7.288 &  & 2.628 &  & 3.171 &  & 0.427 &  & 7.116 &  & 2.595 &  & 1.661 &  & 0.425 \\
			\textbf{PDFuse}~\cite{Cao2025PDFuse} &  & \emph{NeurIPS'25} &  & 6.484 &  & 1.999 &  & 1.176 &  & 0.244 &  & \cellcolor[HTML]{E6E6FA}7.478 &  & 2.573 &  & 2.533 &  & 0.477 &  & 6.922 &  & 2.501 &  & 1.714 &  & 0.381 \\
			\textbf{OmniFuse}~\cite{Zhang2025OmniFuse} &  & \emph{TPAMI'25} &  & 6.889 &  & 2.153 &  & 1.337 &  & 0.407 &  & 7.227 &  & 2.446 &  & 2.401 &  & 0.523 &  & 6.759 &  & 2.541 &  & 1.799 &  & 0.476 \\
			\textbf{DSPFusion}~\cite{Tang2026DSPFusion} &  & \emph{TIP'26} &  & 7.092 &  & 2.308 &  & 1.482 &  & 0.384 &  & \cellcolor[HTML]{FFC7CE}\textbf{7.526} &  & 2.514 &  & \cellcolor[HTML]{E6E6FA}3.269 &  & 0.637 &  & 6.890 &  & 2.638 &  & 2.019 &  & 0.504 \\
			\textbf{ReCoFuse}~\cite{Zhang2026ReCoFuse} &  & \emph{CVPR'26} &  & 6.999 &  & 2.156 &  & 1.357 &  & 0.420 &  & 7.327 &  & \cellcolor[HTML]{E6E6FA}2.842 &  & 3.033 &  & 0.626 &  & 7.100 &  & 2.507 &  & 1.649 &  & 0.552 \\
			\textbf{DiTFuse}~\cite{Li2026DiTFuse} &  & \emph{TPAMI'26} &  & 6.414 &  & 2.157 &  & 1.702 &  & 0.362 &  & 7.123 &  & 2.644 &  & 3.154 &  & 0.671 &  & 6.873 &  & 2.427 &  & 2.056 &  & 0.559 \\
			\textbf{DuS-DiFuse}~\cite{Cao2026DuSDiFuse} &  & \emph{TPAMI'26} &  & \cellcolor[HTML]{E6E6FA}7.152 &  & \cellcolor[HTML]{E6E6FA}2.367 &  & \cellcolor[HTML]{FFC7CE}\textbf{1.919} &  & \cellcolor[HTML]{E6E6FA}0.511 &  & 7.370 &  & 2.539 &  & 2.237 &  & \cellcolor[HTML]{E6E6FA}0.717 &  & \cellcolor[HTML]{E6E6FA}7.186 &  & 2.511 &  & \cellcolor[HTML]{E6E6FA}2.237 &  & \cellcolor[HTML]{FFC7CE}\textbf{0.612} \\
			\midrule
			\textbf{G2Fusion}~\cite{Zhang2026G2Fusion} &  & \emph{NeurIPS'26} &  & 6.539 &  & 2.156 &  & 1.470 &  & 0.362 &  & 7.251 &  & 2.389 &  & 2.062 &  & 0.451 &  & 6.706 &  & 2.347 &  & 1.567 &  & 0.396 \\
			\textbf{GenRF (Ours)} &  & -- &  & \cellcolor[HTML]{FFC7CE}\textbf{7.321} &  & \cellcolor[HTML]{FFC7CE}\textbf{2.397} &  & \cellcolor[HTML]{E6E6FA}1.754 &  & \cellcolor[HTML]{FFC7CE}\textbf{0.560} &  & 7.363 &  & \cellcolor[HTML]{FFC7CE}\textbf{2.847} &  & \cellcolor[HTML]{FFC7CE}\textbf{3.385} &  & \cellcolor[HTML]{FFC7CE}\textbf{0.761} &  & \cellcolor[HTML]{FFC7CE}\textbf{7.388} &  & \cellcolor[HTML]{FFC7CE}\textbf{2.757} &  & \cellcolor[HTML]{FFC7CE}\textbf{2.268} &  & \cellcolor[HTML]{E6E6FA}0.585 \\
			\bottomrule
		\end{tabular}%
	}
	\vspace{-0.05in}
\end{table*}

\subsection{Registration Comparison}
Real Scene and DVTOD contain infrared and visible image pairs with misalignment arising from real acquisition. Since ground-truth alignment is unavailable, we assess registration qualitatively. Figure~\ref{fig:registration-comparison} presents checkerboard visualizations for multi-modal matching methods and fused images of joint registration and fusion methods.

In the first Real Scene case, large depth variation prevents the homographies fitted from MINIMA and AnyMatch correspondences from aligning the foreground and background simultaneously, leaving the foreground pedestrian displaced. Joint registration and fusion methods also introduce artifacts in the highlighted regions, particularly FusionRegister, URMIF, and G2Fusion. With the larger field-of-view mismatch in the second case, these methods exhibit pronounced distortions. MINIMA and AnyMatch provide approximate alignment but retain local offsets in highlighted regions. In contrast, GenRF aggregates complementary information through cross-attention without an explicit warp, enhancing dark regions while preserving coherent pedestrian contours and background structures in both cases, without visible ghosting.

In the first DVTOD example, temporal mismatch further complicates alignment, causing alternatives to introduce artifacts around the moving pedestrian. The final example also reveals ghosting or local offsets around the target and background structures. GenRF retains coherent target contours and finer background details in both examples despite the challenging misalignment.
\begin{figure}[t]
	\centering
	\includegraphics[width=\linewidth]{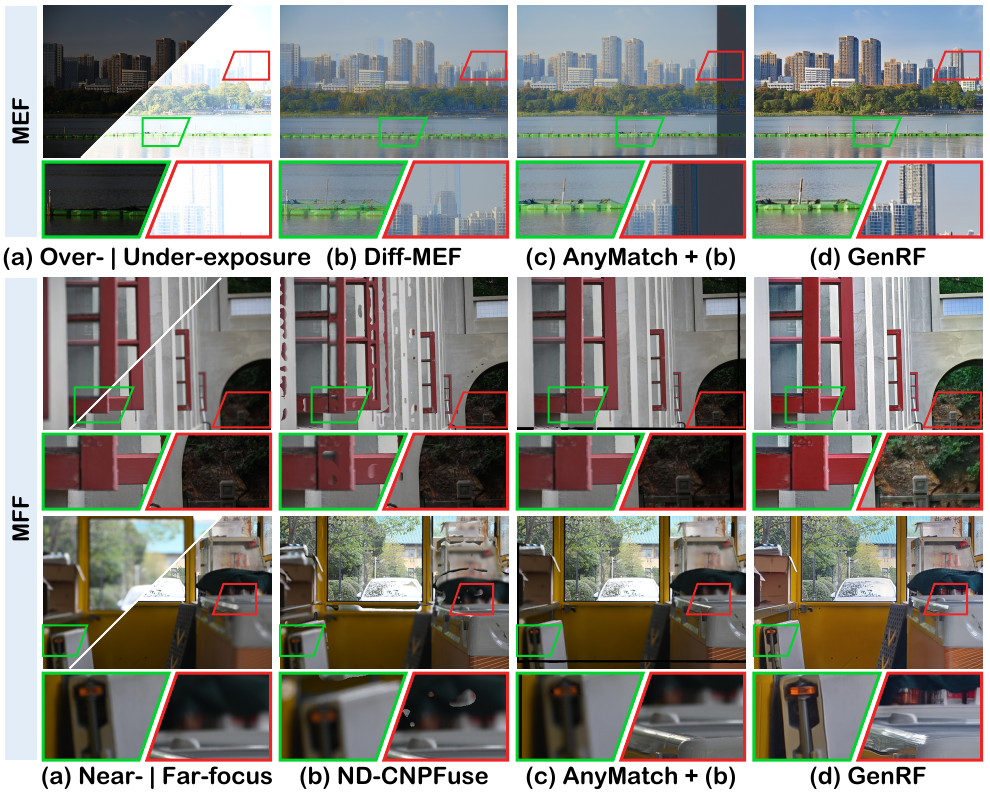}
	\vspace{-0.23in}
	\caption{Qualitative results of multi-exposure and multi-focus image fusion on real misaligned captures.}
	\vspace{-0.3in}
	\label{fig:mef-mff}
\end{figure}

\subsection{Restoration and Fusion Comparison}
Given its overall registration quality, AnyMatch is used to pre-register inputs for restoration and fusion methods. Figure~\ref{fig:real-fusion-comparison} compares fusion results on Real Scene and DVTOD under real-world degradations. In the first Real Scene example, GenRF produces vivid colors and natural transitions between overlapping and non-overlapping regions. It also infers the thermal appearance of the rear wheels where infrared observations are missing. In the second example, residual registration errors still introduce artifacts in alternatives, whereas GenRF produces a brighter scene with coherent structures. In the first DVTOD case, GenRF preserves the inscriptions without interference from strong thermal responses. In the final case, GenRF produces clearer smoke-obscured targets and sharper window details. Overall, pretrained FLUX priors support degradation recovery and missing-content inference, yielding natural colors while suppressing irrelevant infrared responses.

Table~\ref{tab:real-scene-dvtod} compares GenRF with joint registration and fusion methods as well as joint restoration and fusion methods. GenRF achieves the highest DeQA, LIQE, and QualiCLIP scores on both datasets, suggesting better alignment with human visual perception and supporting qualitative analysis. Its competitive EN also reflects substantial information content in fusion results. Table~\ref{tab:msrs-llvip-m3fd} shows consistent advantages across MSRS, LLVIP, and M3FD. Qualitative results are illustrated in Suppl. Material.
\begin{table}[t]
	\centering
	\caption{Quantitative object detection comparison on DVTOD.}    
	\vspace{-0.05in}
	\label{tab:dvtod-detection}
	\setlength{\TableMethodGap}{6pt}
	\setlength{\TableMetricGap}{6pt}
	\setlength{\tabcolsep}{2pt}
	\renewcommand{\arraystretch}{0.98}
	\small
	\resizebox{0.98\linewidth}{!}{%
		\begin{tabular}{@{}l@{}p{\TableMethodGap}@{}c@{}p{\TableMetricGap}@{}c@{}p{\TableMetricGap}@{}c@{}p{\TableMetricGap}@{}c@{}p{\TableMetricGap}@{}c@{}}
			\toprule
			\textbf{Methods} & & Prec.~$\uparrow$ & & Recall~$\uparrow$ & & AP$_{50}$~$\uparrow$ & & AP$_{75}$~$\uparrow$ & & mAP$_{50:95}$~$\uparrow$ \\
			\midrule
			\textbf{Text-IF}~\cite{Yi2024Text-IF} & & \cellcolor[HTML]{FFC7CE}\textbf{0.801} & & 0.637 & & \cellcolor[HTML]{E6E6FA}0.717 & & 0.502 & & 0.480 \\
			\textbf{TG-ECNet}~\cite{Sun2025TGECNet} & & 0.772 & & 0.598 & & 0.680 & & 0.484 & & 0.458 \\
			\textbf{PDFuse}~\cite{Cao2025PDFuse} & & 0.783 & & 0.599 & & 0.682 & & 0.472 & & 0.448 \\
			\textbf{OmniFuse}~\cite{Zhang2025OmniFuse} & & 0.758 & & 0.565 & & 0.637 & & 0.458 & & 0.429 \\
			\textbf{DSPFusion}~\cite{Tang2026DSPFusion} & & 0.780 & & 0.605 & & 0.686 & & 0.465 & & 0.447 \\
			\textbf{ReCoFuse}~\cite{Zhang2026ReCoFuse} & & 0.768 & & 0.580 & & 0.660 & & 0.445 & & 0.429 \\
			\textbf{DiTFuse}~\cite{Li2026DiTFuse} & & 0.781 & & \cellcolor[HTML]{E6E6FA}0.647 & & 0.711 & & \cellcolor[HTML]{E6E6FA}0.537 & & \cellcolor[HTML]{E6E6FA}0.494 \\
			\textbf{DuS-DiFuse}~\cite{Cao2026DuSDiFuse} & & 0.768 & & 0.637 & & 0.713 & & 0.492 & & 0.476 \\
			\midrule
			\textbf{G2Fusion}~\cite{Zhang2026G2Fusion} & & 0.644 & & 0.436 & & 0.457 & & 0.236 & & 0.258 \\
			\textbf{GenRF (Ours)} & & \cellcolor[HTML]{E6E6FA}0.796 & & \cellcolor[HTML]{FFC7CE}\textbf{0.694} & & \cellcolor[HTML]{FFC7CE}\textbf{0.749} & & \cellcolor[HTML]{FFC7CE}\textbf{0.564} & & \cellcolor[HTML]{FFC7CE}\textbf{0.524} \\
			\bottomrule
		\end{tabular}%
	}
	\vspace{-0.15in}
\end{table}

\subsection{Generalization to Multi-exposure and Multi-focus Image Fusion}
We further evaluate GenRF on misaligned multi-exposure and multi-focus image pairs captured in real scenes. As shown in Fig.~\ref{fig:mef-mff}, misalignment introduces visible fusion artifacts in Diff-MEF~\cite{Xu2026DiffMEF} and causes ND-CNPFuse~\cite{Li2026NDCNPFuse} to misidentify in-focus regions, leaving holes in fused images. Therefore, these methods usually require pre-registration methods, such as AnyMatch, to obtain aligned inputs. Notably, GenRF directly fuses misaligned inputs without task-specific fine-tuning and produces well-exposed images with more natural colors and clearer details than the alternatives with pre-registration. Particularly, in the MEF task, the pretrained FLUX prior helps GenRF enhance local contrast while maintaining natural colors. In the MFF task, GenRF combines details from different focus planes into an all-in-focus image while adaptively brightening shadows to reveal details in the middle-row red box. In the bottom-row green box, GenRF renders sharper details than either input, suggesting GenRF can reduce residual blur during fusion. More results and analysis are presented in Suppl. Material.

\subsection{Object Detection under Misalignment}
Detection is evaluated on DVTOD with a pretrained YOLO-11s detector~\cite{Jocher2024YOLO11}. Since the original labels are annotated in infrared coordinates, we re-annotate the dataset in the visible coordinate frame to match the geometry of fusion results. Table~\ref{tab:dvtod-detection} reports the quantitative results. GenRF achieves the highest recall and AP scores, with an mAP$_{50:95}$ of 0.524 compared with 0.494 for the runner-up, DiTFuse. Qualitative results are provided in Suppl. Material.

\subsection{Ablation Study}
\begin{figure}[t]
    \centering
    \includegraphics[width=0.49\textwidth]{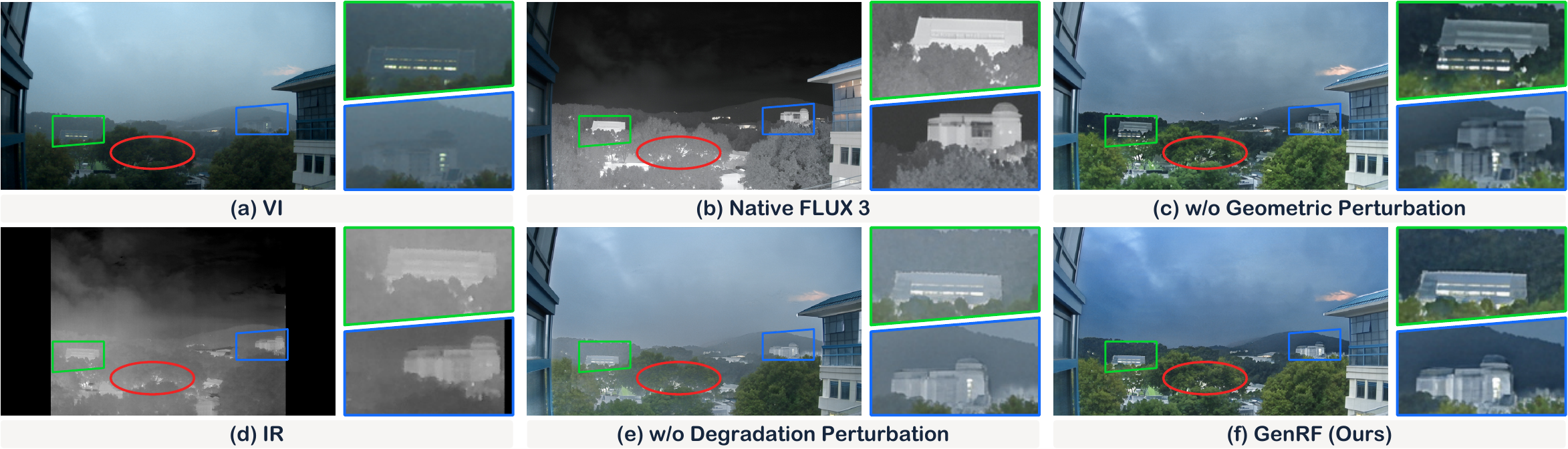}
    \vspace{-0.2in}
    \caption{Qualitative ablation results on Real Scene.}
    \vspace{-0.1in}
    \label{fig:ablation-qualitative}
\end{figure}

\begin{figure}[t]
    \centering
    \includegraphics[width=0.49\textwidth]{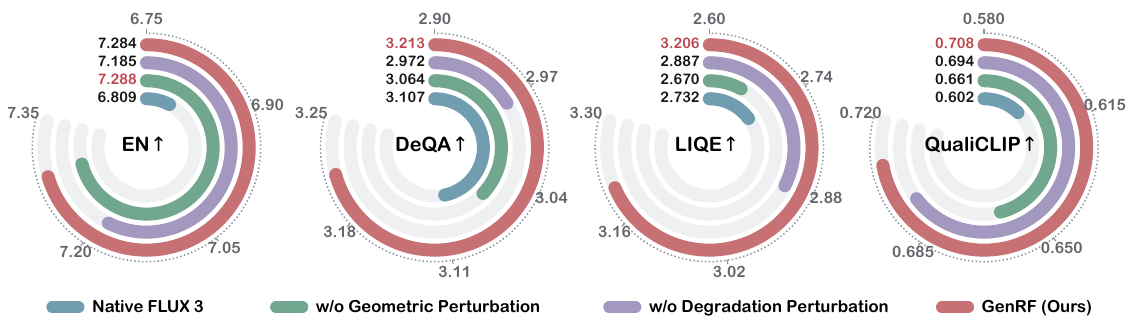}
    \vspace{-0.2in}
    \caption{Quantitative ablation results on Real Scene.}
    \vspace{-0.1in}
    \label{fig:ablation-quantitative}
\end{figure}

GenRF aims to improve the ability of a pretrained generative model to integrate misaligned and degraded observations via LoRA adaptation with geometric and degradation perturbations. As shown in Fig.~\ref{fig:ablation-qualitative}~(b), pretrained FLUX~3 enhances the infrared appearance in its native geometry and uses visible content to fill gaps in infrared coverage, indicating limited cross-modal information aggregation. When fine-tuning omits geometric perturbations, attention still favors matching coordinates, producing spurious contours around displaced salient objects, as shown in the green and blue crops of Fig.~\ref{fig:ablation-qualitative}~(c). Without degradation perturbations during fine-tuning, the fusion result retains visible-image haze and redundant thermal responses, obscuring scene details in the red region of Fig.~\ref{fig:ablation-qualitative}~(e). Together, the two perturbations enable GenRF to aggregate complementary cues across misaligned positions and suppress degradations, revealing clearer scene structures and more distinct objects in Fig.~\ref{fig:ablation-qualitative}~(f). Figure~\ref{fig:ablation-quantitative} corroborates these findings, with GenRF leading in DeQA, LIQE, and QualiCLIP.

\section{Conclusion}
This work revisits misaligned and degraded image fusion under a generative paradigm and proposes \textbf{GenRF}. Built on pretrained FLUX with LoRA, GenRF utilizes cross-attention to aggregate complementary information across misaligned observations, while generative priors support degradation recovery and inference of missing cross-modal content. Online geometric and degradation perturbations improve robustness to diverse input conditions. Experiments on real captures and public datasets demonstrate improved geometric consistency, scene visibility, and fusion quality without separate registration or restoration preprocessing, with benefits for object detection as well as generalization to multi-exposure and multi-focus fusion.

\clearpage
\setcounter{section}{0}
\setcounter{equation}{0}
\setcounter{figure}{0}
\setcounter{table}{0}
\renewcommand{\thesection}{\Alph{section}}
\renewcommand{\theequation}{S\arabic{equation}}
\renewcommand{\thefigure}{S\arabic{figure}}
\renewcommand{\thetable}{S\arabic{table}}
\renewcommand{\theHsection}{supp.\arabic{section}}
\renewcommand{\theHequation}{supp.\arabic{equation}}
\renewcommand{\theHfigure}{supp.\arabic{figure}}
\renewcommand{\theHtable}{supp.\arabic{table}}
\maketitlesupplementary

\raggedbottom

This supplementary material details the geometric perturbations, degradation models, and fusion instructions used in GenRF. It describes the Real Scene acquisition setups, presents fusion results across devices, and analyzes cross-modal attention under simulated misalignment. It also presents qualitative fusion results on MSRS, LLVIP, and M3FD, as well as results for multi-exposure fusion, multi-focus fusion, and downstream object detection. It concludes with a discussion of computational cost.

\section{Geometric Perturbations}
\label{sec:supp-geometry}
Given aligned reference images $I_{ir}^{\mathrm{ref}}$ and $I_{vi}^{\mathrm{ref}}$, we use DuS-DiFuse~\cite{supp-Cao2026DuSDiFuse} to generate the desired fused image $I_f^\star$, which preserves the visible reference geometry. We then geometrically perturb only the infrared image, leaving $I_{vi}^{\mathrm{ref}}$ and $I_f^\star$ fixed. The infrared condition is resampled as follows:
\begin{equation}
\mathcal W_\phi(I_{ir}^{\mathrm{ref}})(\mathbf p)
=\mathcal S\!\left(I_{ir}^{\mathrm{ref}},g_\phi(\mathbf p)+\delta_\phi(\mathbf p)\right),
\label{eq:supp-warp}
\end{equation}
where $\mathbf p\in[-1,1]^2$ is a normalized output-grid location and $\mathcal S$ performs bilinear resampling. When infrared degradation is applied, the same warp is applied to the degraded image.

\subsection{Global transformations.} 
To simulate overall geometric discrepancies caused by different imaging configurations, we sample rotation, translation, anisotropic scaling, and optional projective distortion:
\begin{equation}
g_\phi(\mathbf p)=
\frac{A_\phi\mathbf p+\mathbf t_\phi}
{1+\mathbf h_\phi^\top\mathbf p},
\label{eq:supp-global}
\end{equation}
where $A_\phi$ combines rotation and scaling, $\mathbf t_\phi$ specifies translation, and $\mathbf h_\phi$ controls the projective component. The transformation becomes affine when $\mathbf h_\phi=\mathbf 0$, while a nonzero $\mathbf h_\phi$ introduces projective distortion.

\subsection{Local displacements.} 
A global transformation alone cannot capture spatially varying misalignment caused by depth-dependent parallax. Therefore, we introduce smooth local displacements by bicubically upsampling a two-channel Gaussian random field from a $7\times9$ grid. Each component is normalized by its spatial standard deviation to control deformation strength. For parallax-oriented perturbations, we additionally introduce a regional offset:
\begin{equation}
\delta_\phi(\mathbf p)=\mathbf a_\phi\odot\bar e_\phi(\mathbf p)
+b_\phi(\mathbf p)\mathbf u_\phi,
\label{eq:supp-elastic}
\end{equation}
where $\bar e_\phi$ is the normalized field and $\mathbf a_\phi$ converts the sampled pixel amplitudes to normalized coordinates. Here, $\odot$ denotes elementwise multiplication. The soft mask $b_\phi$ restricts the additional offset $\mathbf u_\phi$ to a local region and is used only for parallax-oriented perturbations.

The sampler leaves 25\% of the image pairs geometrically aligned. The remaining probabilities are $0.35$ for mild global--elastic perturbations, $0.25$ for camera-view perturbations, and $0.15$ for parallax-oriented perturbations. Rotation angles and translation offsets are sampled with maximum magnitudes of $10^\circ$ and $18\%$ of the corresponding image dimension, respectively, and the principal scale range is $[0.75,1.28]$. Elastic amplitudes range from 6 to 12 pixels at the training resolution, with an additional aspect-ratio variation in camera-view perturbations. Parameters are resampled on each visit, and out-of-bounds coordinates are filled with black to represent uncovered infrared regions.

\section{Degradation Construction}
\label{sec:supp-degradation}
We apply modality-specific degradation operators $\mathcal D_m$ to the high-quality source images from DSPFusion~\cite{supp-Tang2026DSPFusion} to construct degraded conditions:
\begin{equation}
\begin{aligned}
\widetilde I_{ir}&=\mathcal W_\phi\!\left(\mathcal D_{ir}(I_{ir}^{\mathrm{ref}};\eta_{ir})\right),\\
\widetilde I_{vi}&=\mathcal D_{vi}(I_{vi}^{\mathrm{ref}};\eta_{vi}).
\end{aligned}
\label{eq:supp-conditions}
\end{equation}
where $m\in\{ir,vi\}$ denotes the modality and $\eta_m$ controls degradation type and strength.
The infrared image is degraded before geometric perturbation, whereas the visible image undergoes degradation only.

\begin{table*}[t]
    \centering
    \caption{Degradation models and parameter ranges for constructing source conditions. }
    \label{tab:supp-operators}
    \label{tab:supp-severity}
    \label{tab:supp-degradations}
    \small
    \renewcommand{\arraystretch}{1.24}
    \setlength{\tabcolsep}{1.5pt}
    \setlength{\extrarowheight}{1.5pt}
    \begin{tabularx}{\textwidth}{l>{\raggedright\arraybackslash}p{0.38\textwidth}>{\raggedright\arraybackslash}Xccc}
    \toprule
    \rowcolor{black!5}
    \textbf{Degradation} & \textbf{Construction} & \textbf{Parameter} & \textbf{Light} & \textbf{Medium} & \textbf{Heavy} \\
    \midrule
    \multicolumn{6}{l}{\textcolor{cvprblue}{\textbf{Visible}}} \\
    \cmidrule(lr){1-6}
    \rowcolor{degvisible!26}
    \textbf{Low-light} & Illumination adjustment via intrinsic decomposition & Intrinsic gain $g_e$ & $0.35$--$0.65$ & $0.10$--$0.30$ & $0.03$--$0.09$ \\
    \cmidrule(lr){1-6}
    \rowcolor{degvisible!23}
    \textbf{Over-exposure} & Illumination adjustment via intrinsic decomposition & Intrinsic gain $g_e$ & $1.3$--$2.0$ & $2.5$--$4.0$ & $5.0$--$8.0$ \\
    \cmidrule(lr){1-6}
    \rowcolor{degvisible!20}
    \textbf{Haze} & Atmospheric scattering with estimated depth & Density $\beta$ & $0.3$--$0.7$ & $0.7$--$1.4$ & $1.4$--$2.2$ \\
    \cmidrule(lr){1-6}
    \rowcolor{degvisible!17}
    &  & Density $\rho_r$ & $0.8$--$1.5$ & $1.5$--$3.0$ & $3.0$--$5.0$ \\
    \rowcolor{degvisible!17}
    &  & Length $\ell_r$ & $7$--$12$ & $7$--$20$ & $7$--$28$ \\
    \rowcolor{degvisible!17}
    \multirow{-3}{*}{\textbf{Rain streaks}} & \multirow{-3}{0.38\textwidth}{Semi-transparent streak compositing} & Opacity $o_r$ & $25$--$45$ & $40$--$65$ & $55$--$85$ \\
    \cmidrule(lr){1-6}
    \rowcolor{degvisible!14}
    \textbf{Raindrops} & Local blur and fisheye remapping & Radius $r_d$ & $8$--$17$ & $9$--$20$ & $10$--$23$ \\
    \cmidrule(lr){1-6}
    \rowcolor{degvisible!11}
    \textbf{JPEG} & JPEG encoding and decoding & Quality $Q$ & $75$--$95$ & $40$--$74$ & $15$--$39$ \\
    \cmidrule(lr){1-6}
    \rowcolor{degvisible!8}
    \textbf{Motion blur} & Convolution with a directional kernel & Length $\ell_m$ & $3$--$5$ & $5$--$9$ & $9$--$15$ \\
    \cmidrule(lr){1-6}
    \multicolumn{6}{l}{\textcolor{cvprblue}{\textbf{Infrared}}} \\
    \cmidrule(lr){1-6}
    \rowcolor{deginfrared!20}
    &  & Compression $\kappa$ & $0.75$--$0.90$ & $0.50$--$0.75$ & $0.30$--$0.50$ \\
    \rowcolor{deginfrared!20}
    \multirow{-2}{*}{\textbf{Low-contrast}} & \multirow{-2}{0.38\textwidth}{Gamma adjustment and contrast compression} & Gamma $\gamma$ & $0.90$--$1.10$ & $0.80$--$1.25$ & $0.70$--$1.40$ \\
    \cmidrule(lr){1-6}
    \rowcolor{deginfrared!10}
    \textbf{Stripes} & Column-dependent offsets shared across rows & Column-offset std. $\sigma_c$ & $0.008$--$0.018$ & $0.018$--$0.035$ & $0.035$--$0.060$ \\
    \cmidrule(lr){1-6}
    \multicolumn{6}{l}{\textcolor{cvprblue}{\textbf{Visible / infrared}}} \\
    \cmidrule(lr){1-6}
    \rowcolor{degshared!18}
    &  & Read-noise std. $\sigma_r$ & $0.004$--$0.012$ & $0.012$--$0.025$ & $0.025$--$0.045$ \\
    \rowcolor{degshared!18}
    \multirow{-2}{*}{\textbf{Noise}} & \multirow{-2}{0.38\textwidth}{Signal-dependent Poisson and Gaussian read noise} & Photon scale $\nu$ & $900$--$1800$ & $300$--$900$ & $100$--$300$ \\
    \cmidrule(lr){1-6}
    \rowcolor{degshared!8}
    \textbf{Gaussian blur} & Convolution with a Gaussian kernel & Radius $r_b$ & $0.4$--$1.0$ & $1.0$--$2.0$ & $2.0$--$3.5$ \\
    \bottomrule
    \end{tabularx}
\end{table*}

\subsection{Degradation Sampling Strategy}
Visible low-light and over-exposure images and infrared low-contrast images provide naturally occurring degradations, while other degradation types are synthesized. For each selected degradation type, we randomly choose an existing degraded image, further degrade that image, or synthesize degradation from a high-quality reference. Infrared low-contrast inputs are taken from the low-contrast images provided in the dataset, with additional corruption preserving their low-contrast characteristics. Synthetic degradations use light, medium, and heavy levels in proportions of $45{:}40{:}15$. Additional corruption uses light and medium levels in proportions of $70{:}30$ to limit severity. 

\subsection{Degradation Models}
All operators use intensities in $[0,1]$, with clipping after corruption. Table~\ref{tab:supp-degradations} summarizes the degradation models and parameter ranges, where the parameters are sampled uniformly. In the following formulations, $I$ denotes the input image to each degradation operator, and $Y(\cdot)$ extracts luminance. Both depth estimation and intrinsic image decomposition take $I_{vi}^{\mathrm{ref}}$ as input.

\textbf{Low-light and over-exposure.} We simulate exposure changes by adjusting illumination while preserving reflectance. Intrinsic v2~\cite{supp-Careaga2024Colorful} estimates reflectance $R$ and shading $L$ from the linearized visible reference $I_{\mathrm{lin}}=(I_{vi}^{\mathrm{ref}})^{2.2}$. With residual $E=I_{\mathrm{lin}}-R\odot L$, the degraded image is constructed as:
\begin{equation}
\begin{aligned}
I'&=\clip\!\left(G\odot(R\odot L+E_-)+kE_+,0,1\right)^{1/2.2},\\
G&=g_eF,
\end{aligned}
\label{eq:supp-exposure}
\end{equation}
where $g_e$ controls overall exposure and $F$ introduces smooth spatial variation based on shading. The gain ranges in Table~\ref{tab:supp-degradations} use $g_e<1$ for low-light and $g_e>1$ for over-exposure. The positive and negative residuals are $E_+=\max(E,0)$ and $E_-=\min(E,0)$, with $k$ controlling the positive-residual gain.

\textbf{Haze.} We simulate haze using the atmospheric scattering model~\cite{supp-Narasimhan2002Atmosphere}, which combines depth-dependent attenuation with atmospheric light:
\begin{equation}
I'(\mathbf p)=t(\mathbf p)I(\mathbf p)
+a\bigl(1-t(\mathbf p)\bigr),\quad
t(\mathbf p)=e^{-\beta d(\mathbf p)},
\label{eq:supp-haze}
\end{equation}
where $t$ denotes transmission, $a\in[0.7,0.95]$ is atmospheric light, and $\beta$ controls haze density. We obtain $d\in[0,1]$ from the visible reference using Depth Anything 3~\cite{supp-Lin2025DepthAnything3}.

\textbf{Rain streaks and raindrops.} We employ a simplified alpha-compositing model~\cite{supp-Garg2006RainStreaks} to simulate rain streaks with random positions and slant:
\begin{equation}
I'(\mathbf p)=\bigl(1-\omega(\mathbf p)\bigr)I(\mathbf p)
+\omega(\mathbf p)C_r,
\label{eq:supp-rain}
\end{equation}
where $C_r$ is the streak color and $\omega(\mathbf p)$ gives the opacity at pixel $\mathbf p$ for a map formed by drawing streaks of vertical length $\ell_r$ and opacity $o_r/255$, with their number determined by the density $\rho_r$. Table~\ref{tab:supp-degradations} gives $\rho_r$ in units of $10^{-4}$ streaks per pixel, $\ell_r$ in pixels, and $o_r$ on a 0--255 scale. Raindrops are simulated using ROLE~\cite{supp-ROLERaindrop}, which applies local blur and fisheye-like remapping with radius $r_d$.

\textbf{JPEG.} We apply JPEG encoding and decoding with quality $Q$ and 4:2:0 chroma subsampling. Lower $Q$ produces stronger quantization and chroma loss.

\textbf{Motion blur.} We simulate motion blur by convolving the input image with a directional kernel:
\begin{equation}
    I'=K*I,\qquad \sum\nolimits_{\mathbf p}K(\mathbf p)=1.
    \label{eq:supp-blur}
\end{equation}
The kernel length $\ell_m$ is measured in pixels and follows Table~\ref{tab:supp-severity}, with motion angles sampled from $[0,\pi)$.

\textbf{Low-contrast.} We model infrared low-contrast by applying gamma adjustment and compressing contrast around the mean luminance:
\begin{equation}
I'=\clip\!\left(\mu_\gamma+\kappa_{\mathrm{eff}}(I^\gamma-\mu_\gamma),0,1\right),
\label{eq:supp-contrast}
\end{equation}
where $\mu_\gamma=\operatorname{mean}(Y(I^\gamma))$. To prevent gamma-induced contrast expansion, we set:
\begin{equation}
\kappa_{\mathrm{eff}}=\kappa \times \min\!\left(1,
\frac{\operatorname{std}(Y(I))}{\operatorname{std}(Y(I^\gamma))}\right).
\label{eq:supp-contrast-guard}
\end{equation}
For constant inputs, $\kappa_{\mathrm{eff}}=\kappa$. Thus, gamma cannot counteract the intended contrast reduction.

\textbf{Stripes.} We simulate vertical stripes by adding an independently sampled brightness offset to each image column:
\begin{equation}
\begin{aligned}
I'(x,y)&=\clip\bigl(I(x,y)+s(x),0,1\bigr),\\
s(x)&\sim\mathcal N(0,\sigma_c^2),
\end{aligned}
\label{eq:supp-stripes}
\end{equation}
where $x$ and $y$ index columns and rows. Offsets are shared across rows and channels, and $\sigma_c$ controls stripe strength.

\textbf{Noise.} We combine signal-dependent shot noise with signal-independent read noise:
\begin{equation}
\begin{aligned}
I'&=\clip(I+\epsilon_s+\epsilon_r,0,1),\\
\epsilon_s&=\operatorname{Poisson}(\nu I)/\nu-I,
\quad\epsilon_r\sim\mathcal N(0,\sigma_r^2).
\end{aligned}
\label{eq:supp-noise}
\end{equation}
Here, $\nu$ controls the photon scale and $\sigma_r$ controls read-noise strength. Smaller $\nu$ increases signal-dependent fluctuations. Visible noise is sampled independently across channels. For infrared inputs, the grayscale signal determines the shot noise, and the same noise realization is shared across channels to avoid artificial color interference.

\textbf{Gaussian blur.} Gaussian blur is simulated using the convolution in Eq.~\eqref{eq:supp-blur}, with radius $r_b$ specified in Table~\ref{tab:supp-severity}.

\subsection{Compound Degradation}
Real captures often exhibit compound degradations, such as low-light accompanied by noise. Therefore, we combine selected degradation operators to construct compound conditions. The permitted combinations include visible low-light with noise, exposure with rain streaks, raindrops, haze, or JPEG, and infrared low-contrast with stripes. 
For an eligible primary type, the probability of selecting a second operator is $0.35$ for synthesis from a reference and $0.25$ for additional corruption of a supplied variant. Operators follow a fixed order: exposure or contrast, weather interference, Gaussian or motion blur, stripes, noise, and JPEG. 

\begin{figure}[t]
    \centering
    \includegraphics[width=0.49\textwidth]{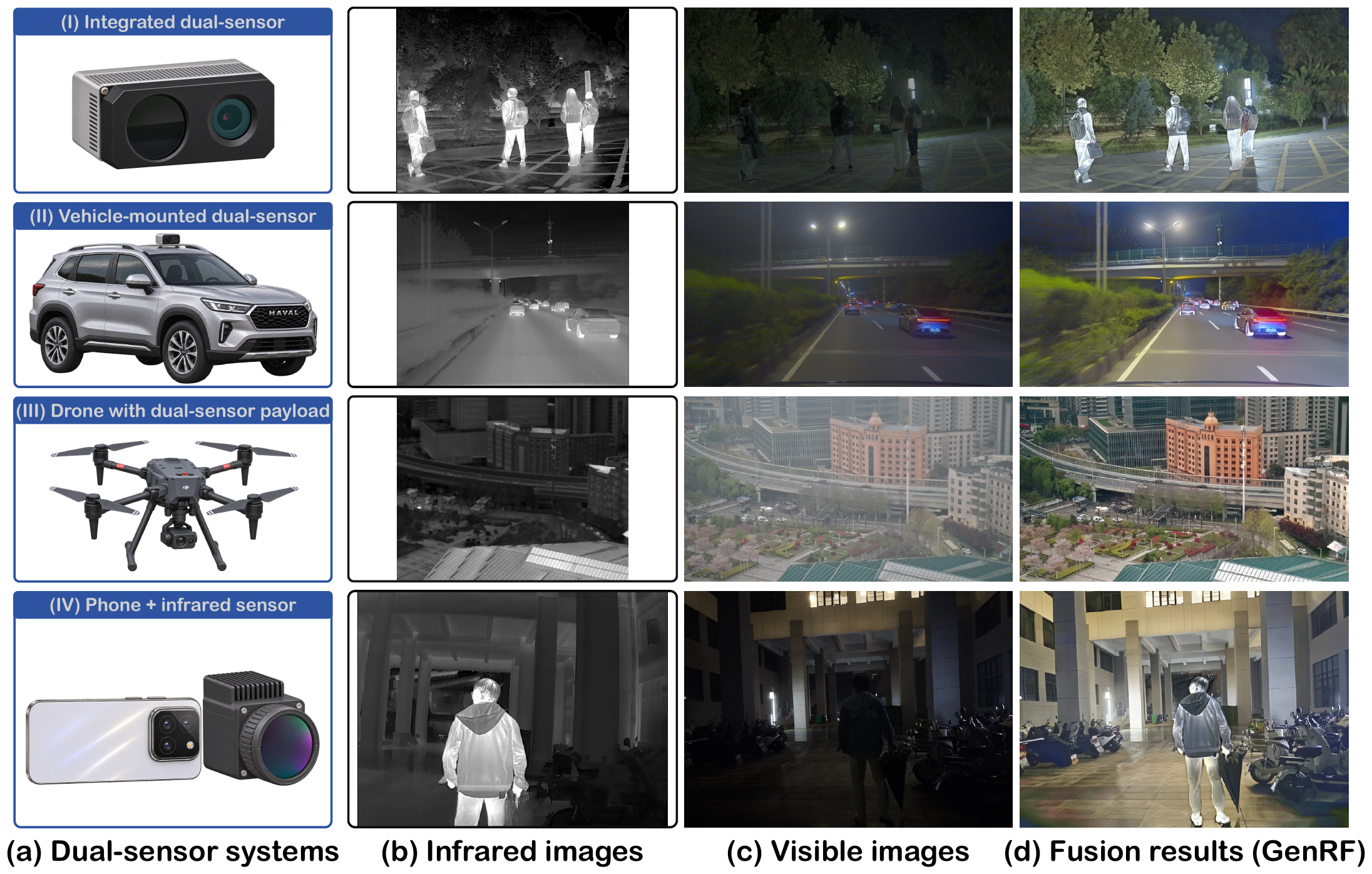}
    \caption{Real-world dual-sensor imaging systems, infrared and visible captures, and fusion results. (I) Our integrated dual-sensor prototype, (II) the vehicle-mounted dual-sensor system used in road tests, (III) a DJI drone carrying a custom dual-sensor payload, and (IV) a phone paired with our custom infrared sensor.}
    \vspace{-0.1in}
    \label{fig:supp-cross-device}
\end{figure}

\begin{figure*}[!t]
\centering
\includegraphics[width=\textwidth]{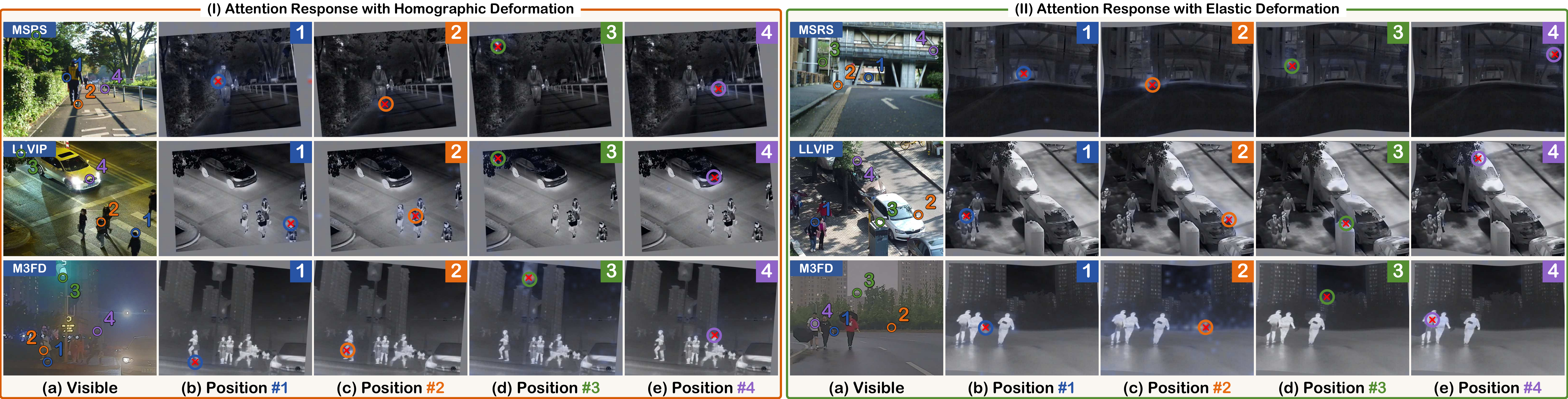}
\caption{Attention responses under simulated infrared deformations: (I) homographic and (II) elastic. Colored $\bm{\circ}$ mark the transformed reference locations of numbered visible queries, while \textcolor{red}{$\bm{\times}$} marks the maximum of each averaged attention map.}
\label{fig:supp-simulated-attention}
\end{figure*}

\section{Fusion Instructions}
\label{sec:supp-instructions}
Training instructions combine common fusion requirements with scene-specific descriptions to guide both task objectives and the desired scene appearance. The common instruction calls for restoring degraded content while preserving scene geometry and useful infrared information:
\begin{tcolorbox}[fusionprompt, title={Common instruction},colback=degvisible!5,colbacktitle=degvisible!13,colframe=degvisible!40]
Fuse the infrared and visible images. Restore degraded visibility, reduce noise, preserve the visible scene geometry and natural colors, and retain useful infrared information. Make source-supported salient targets clear while preserving internal shading and fine texture. Avoid duplicated contours, halos, fabricated objects, and excessive sharpening.
\end{tcolorbox}
To specify the content and appearance of each desired fused image, we append a scene-specific description of the corresponding DuS-DiFuse result to the common instruction, using ChatGPT vision to describe the scene layout, exposure, colors, salient targets, and textures. Each description remains fixed under source perturbations to maintain consistency with the desired fused image. An example is:
\begin{tcolorbox}[fusionprompt,title={Scene-specific description},colback=deginfrared!5,colbacktitle=deginfrared!13,colframe=deginfrared!40]
The scene is an elevated nighttime sidewalk with a running pedestrian and parked cars. Maintain clear separation between the runner and the surroundings, brighter sidewalk midtones, and warm color tones throughout the background.
\end{tcolorbox}
For infrared and visible image fusion, we use the same instruction for all test pairs:
\begin{tcolorbox}[fusionprompt,title={Test instruction},colback=degshared!5,colbacktitle=degshared!13,colframe=degshared!40]
Fuse the infrared and visible images. Restore degraded visibility, preserve the visible scene geometry and natural colors, and make salient infrared targets clear without ghosts or halos.
\end{tcolorbox}

\begin{figure*}[t]
    \centering
    \includegraphics[width=\textwidth]{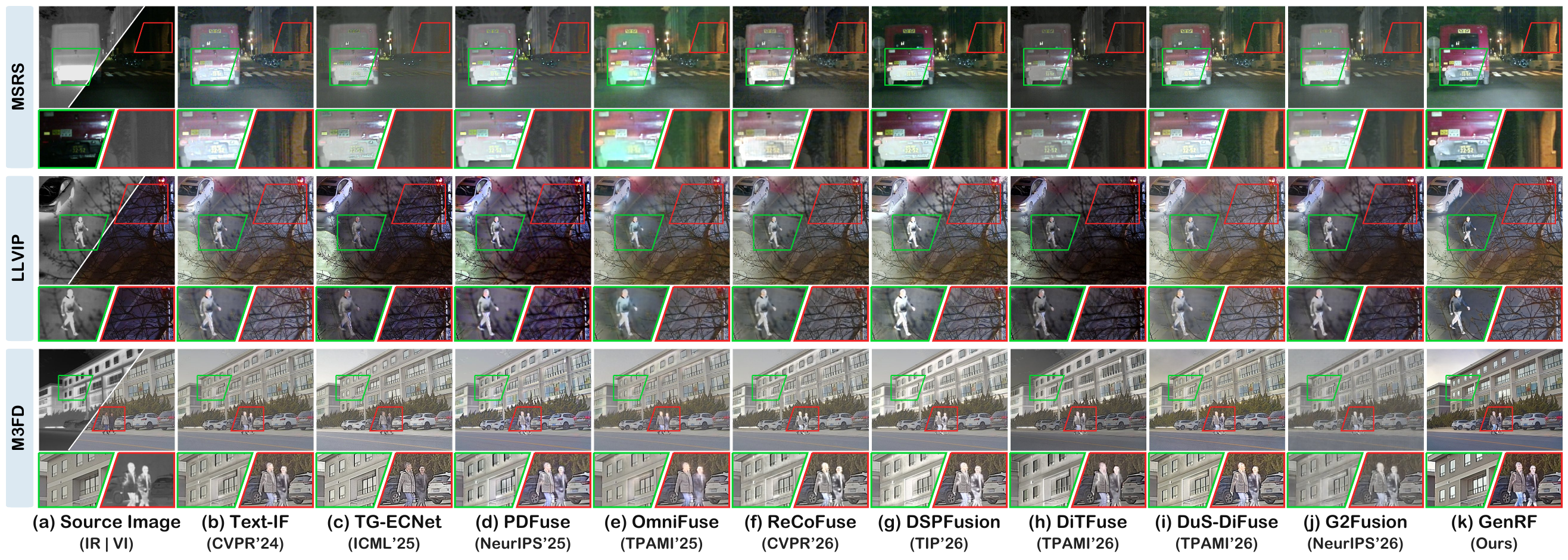}
    \caption{Qualitative fusion comparison on MSRS, LLVIP, and M3FD.}
    \label{fig:supp-standard-fusion}
    % \vspace{-0.1in}
\end{figure*}

\section{Real Scene Acquisition and Fusion Results}
\label{sec:supp-cross-device}
To evaluate GenRF under real acquisition conditions, we collected the Real Scene image pairs using multiple dual-sensor imaging systems. Figure~\ref{fig:supp-cross-device} shows representative infrared and visible pairs and their fused images for four configurations: (I) our integrated dual-sensor prototype, (II) a vehicle-mounted dual-sensor system, (III) a DJI drone with a custom dual-sensor payload, and (IV) a phone paired with our custom infrared sensor. The configurations cover different viewing distances, sensor placements, and lighting conditions, ranging from nearby pedestrians in low light to a hazy aerial view. Specifically, the phone and infrared camera are separate devices that capture images \textit{asynchronously} from various viewpoints and with different fields of view.

In the integrated-sensor and phone-based examples, the pedestrians have stronger contrast in the infrared images, while the visible images retain scene color but reveal little detail in dark regions. GenRF enhances both low-light scenes while preserving clear pedestrian contours, revealing trees and pavement in the integrated-sensor example as well as the corridor and parked motorcycles in the phone-based example. Notably, in the phone-based case, GenRF preserves the observed thermal cues and \textit{leverages pretrained generative priors to infer details of the pedestrian's legs beyond the infrared field of view}. In the vehicle-mounted example, the infrared image highlights vehicles ahead, while the visible image captures their light colors and the surrounding road. GenRF preserves these colors and improves the visibility of the lane markings and roadside vegetation. In the aerial example, haze reduces contrast in the visible view, especially around distant buildings and roads. GenRF produces clearer building facades and road structures while retaining natural scene colors. These examples show that GenRF integrates complementary infrared and visible cues and improves scene visibility across acquisition systems.

\section{Attention Responses under Misalignment}
\label{sec:supp-attention}
Figure~3 in the main paper presents attention responses on misaligned infrared and visible image pairs captured in real scenes. To assess how accurately these responses localize corresponding infrared content, we apply simulated homographic and elastic deformations to the infrared images of aligned pairs from MSRS, LLVIP, and M3FD while keeping the visible images fixed. For each visible query, the simulated deformation gives the expected position of the corresponding scene point in the warped infrared image, allowing us to compare this reference position with the attention peak in Fig.~\ref{fig:supp-simulated-attention}.

Following ZeroCo~\cite{supp-An2025ZeroCo}, we examine the visible-to-infrared interaction by extracting the corresponding submatrix from the joint attention weights computed after query/key normalization and rotary positional encoding. For each visible query $i$, the averaged infrared response is:
\begin{equation}
\bar{\mathcal A}_{vi,ir}(i,j)=\frac{1}{|\Lambda||\mathcal H|}
\sum_{\ell\in\Lambda}\sum_{h\in\mathcal H}
\mathcal A_{vi,ir}^{(\ell,h)}(i,j),
\label{eq:supp-attention-map}
\end{equation}
where $j$ indexes infrared tokens, $\Lambda=\{0,6,13,20,27\}$ indexes the selected transformer blocks, and $\mathcal H$ contains all 24 heads. Responses are recorded at denoising step 10. For these simulations, we extract the visible-to-infrared weights from the joint attention probabilities, retaining the softmax normalization over the full token sequence. Circles of the same color link each visible-image query to its corresponding reference location in the deformed infrared image. Red crosses indicate the maxima of the averaged attention maps, showing their agreement with the reference locations.

\textbf{Homographic deformation.} In Fig.~\ref{fig:supp-simulated-attention}~(I), MSRS queries on the pedestrian and vegetation produce localized peaks near their transformed infrared reference locations. The LLVIP example contains densely distributed pedestrians with similar infrared appearances. Despite their spatial proximity, queries on different pedestrians yield distinct peaks in their respective infrared regions, while vehicle queries respond to the corresponding vehicle regions. Thus, the responses distinguish neighboring subjects instead of collapsing onto a single pedestrian. The nighttime M3FD case further shows localized responses for both foreground and background queries despite cross-modal appearance differences.

\textbf{Elastic deformation.} Elastic deformation poses a greater challenge because displacements vary across the image, as illustrated in Fig.~\ref{fig:supp-simulated-attention}~(II). It is worth noting that attention peaks generally remain close to the corresponding infrared reference locations. In MSRS, this agreement holds for queries on both the road and the overhead structure. The LLVIP responses distinguish pedestrians from vehicles, and the M3FD responses localize ground and background regions even in the presence of bright infrared pedestrians. Together with the real examples in the main paper, these results indicate that cross-attention can associate corresponding scene content across misaligned positions and support complementary information aggregation without explicit registration.

\section{Additional Fusion Results and Analysis}
\label{sec:supp-fusion}
Figure~\ref{fig:supp-standard-fusion} complements the quantitative comparisons on MSRS, LLVIP, and M3FD, showing that GenRF restores degraded content alongside handling misalignment within the fusion process. In the extremely dark MSRS example, alternatives retain dark surroundings or overexpose the bus rear. GenRF reveals trees and road details while preserving the bus appearance and nighttime colors. The LLVIP example contains residual misalignment around tree branches, which produces duplicated contours in several alternatives. GenRF preserves branch structures without visible ghosting and retains the pedestrian's clothing details. In M3FD, the visible input has a hazy appearance, whereas GenRF produces clearer building facades, windows, and pedestrians with natural colors.

\begin{figure*}[t]
    \centering
    \includegraphics[width=\textwidth]{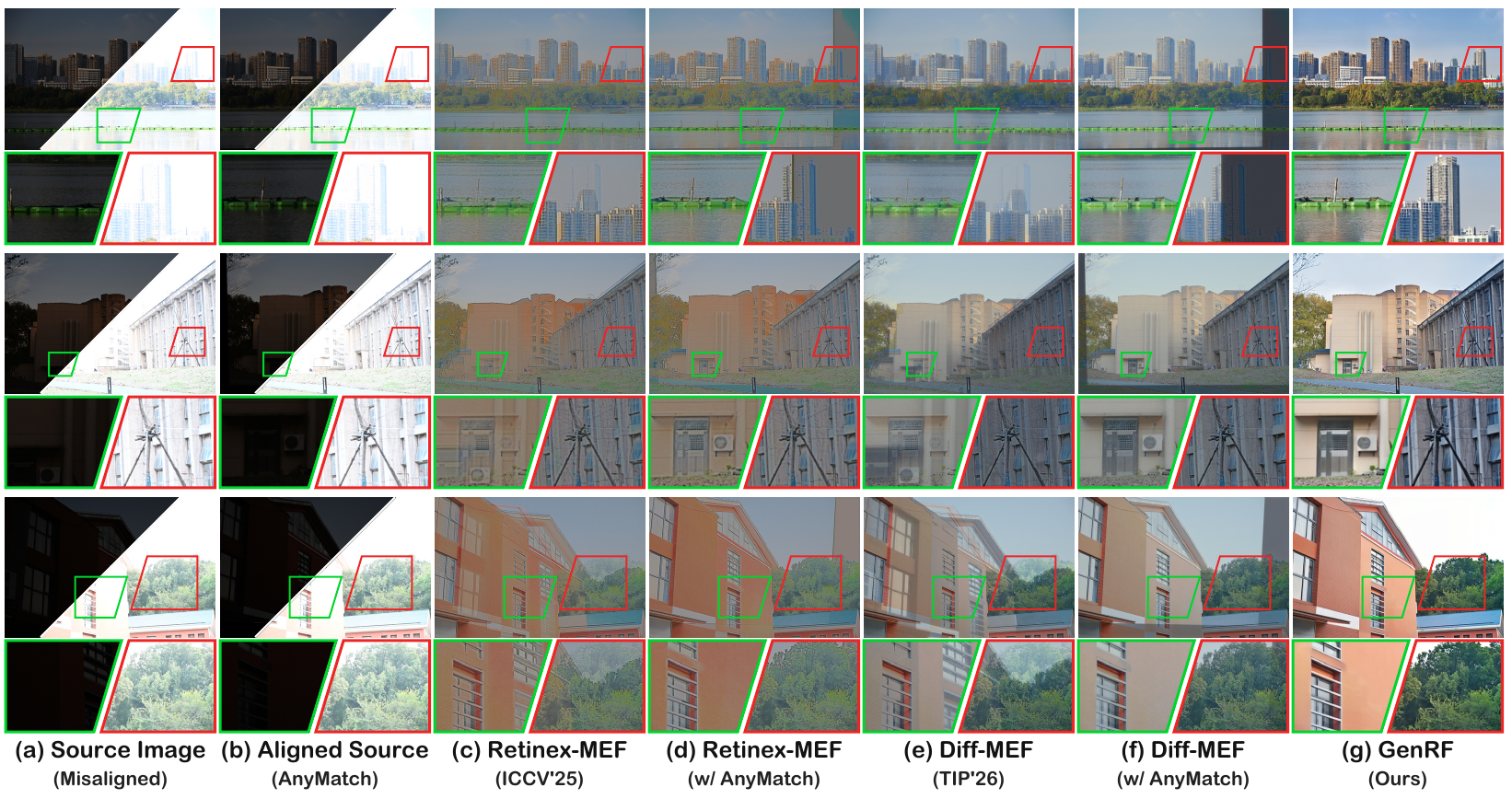}
    \vspace{-0.25in}
    \caption{Qualitative comparison of multi-exposure image fusion on misaligned image pairs captured in real scenes. (a) shows the misaligned inputs. In (b), AnyMatch aligns the underexposed image to the overexposed image. (c) and (e) show the results of Retinex-MEF and Diff-MEF on the misaligned inputs. (d) and (f) show the results of Retinex-MEF and Diff-MEF with AnyMatch pre-registration. (g) shows GenRF applied directly to the misaligned multi-exposure images without fine-tuning.}
    \label{fig:supp-mef}
\end{figure*}

Similar to GenRF, G2Fusion~\cite{supp-Zhang2026G2Fusion} jointly addresses misalignment and degradation through generative fusion. However, its MSRS result overexposes the highlighted bus region and obscures license-plate details, while its M3FD result exhibits blur around salient targets and buildings. Since G2Fusion aligns its inputs using dense flow, residual misalignment can carry over into the fused image. In contrast, GenRF uses cross-attention to retrieve complementary cues across misaligned positions, yielding sharper object contours and clearer building details in the M3FD example. Furthermore, G2Fusion learns a task-specific generative fusion model, while GenRF inherits image priors from large-scale FLUX pretraining. In summary, cross-attention and pretrained generative priors jointly help GenRF preserve scene structures and restore degraded content.

Notably, GenRF surpasses its teacher, DuS-DiFuse, in nighttime background visibility and building clarity. DuS-DiFuse generates the training targets from high-quality aligned source images, while GenRF learns from perturbed versions of those sources. These gains are consistent with the complementary roles of pretrained FLUX priors and geometric perturbation training. On the one hand, generative priors from pretrained FLUX promote more natural colors and balanced exposure, allowing the fusion appearance to improve beyond that provided by teacher supervision. On the other hand, geometric perturbation training encourages cross-attention to aggregate complementary information across displaced positions, improving tolerance to parallax. These capabilities jointly contribute to improved fusion quality on real captures, as shown in the main paper.

\section{Generalization to Multi-exposure and Multi-focus Image Fusion}
\label{sec:supp-generalization}
We extend the evaluation to real misaligned multi-exposure and multi-focus images and compare GenRF with Retinex-MEF~\cite{supp-Bai2025RetinexMEF} and Diff-MEF~\cite{supp-Xu2026DiffMEF} for multi-exposure image fusion, and with UHD-MFF~\cite{supp-Zhang2026UHDMFF} and ND-CNPFuse~\cite{supp-Li2026NDCNPFuse} for multi-focus image fusion. For each task, we test the alternative methods on the original misaligned pairs and on pairs pre-registered by AnyMatch. GenRF~directly fuses the misaligned pairs without task-specific fine-tuning. 

\begin{figure*}[t]
    \centering
    \includegraphics[width=\textwidth]{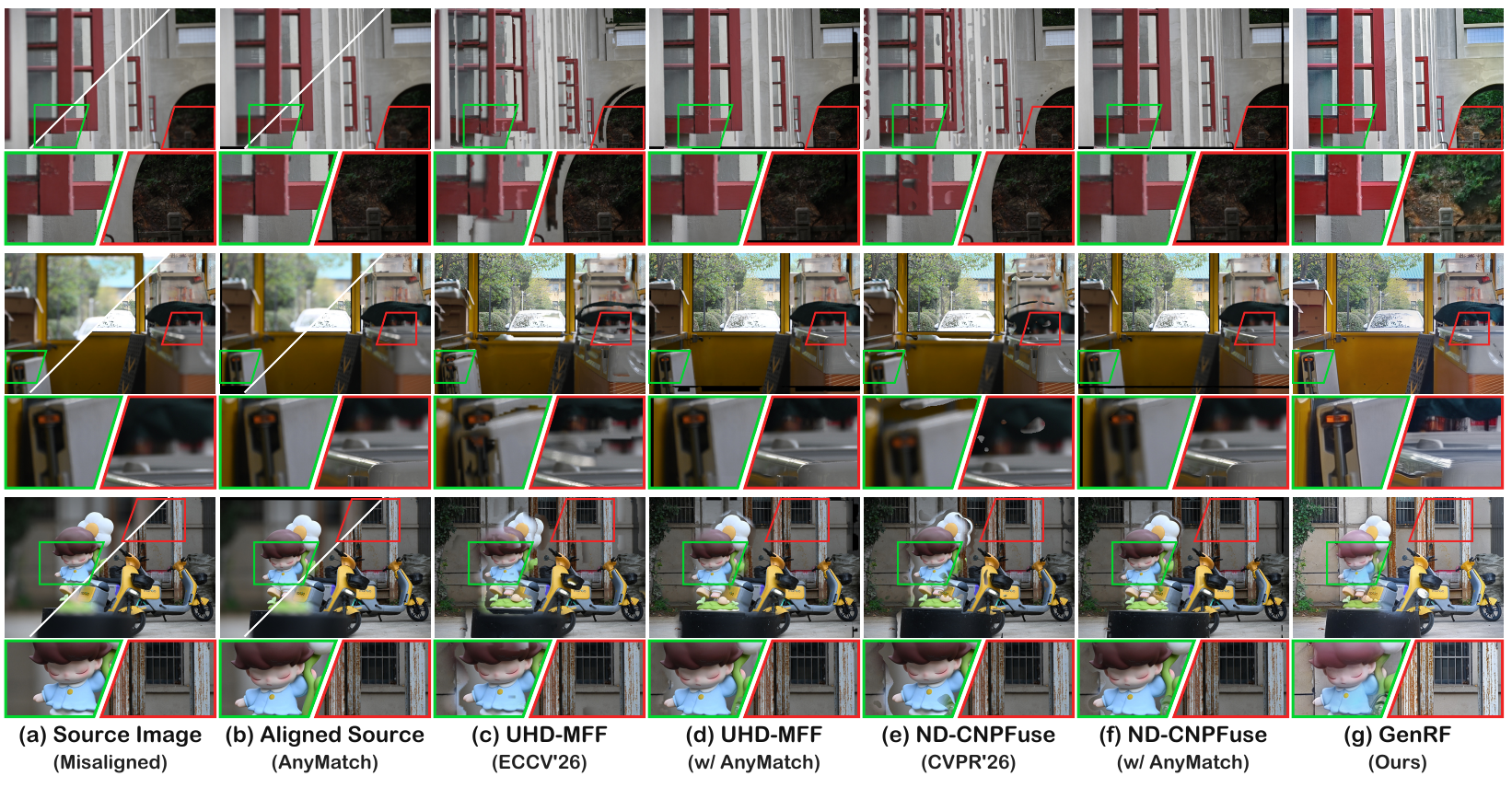}
    \vspace{-0.25in}
    \caption{Qualitative comparison of multi-focus image fusion on misaligned image pairs captured in real scenes. (a) shows the misaligned inputs, with crops from the in-focus regions of the respective source images. In (b), AnyMatch aligns the near-focus image to the far-focus image. (c) and (e) show results from UHD-MFF and ND-CNPFuse on the misaligned inputs. (d) and (f) show results from UHD-MFF and ND-CNPFuse with AnyMatch pre-registration. (g) shows GenRF applied directly to misaligned multi-focus images without fine-tuning.}
    \label{fig:supp-mff}
\end{figure*}

\textbf{Multi-exposure image fusion.} We use the test instruction: \inlineprompt{Fuse the two differently exposed images of the same scene into a naturally exposed RGB image. Preserve scene geometry, realistic colors, and fine details in both shadows and highlights. Avoid ghosting, halos, clipped highlights, and invented details}, for multi-exposure image fusion. As shown in Fig.~\ref{fig:supp-mef}, directly fusing misaligned inputs introduces visible artifacts in both Retinex-MEF and Diff-MEF, particularly along building edges. In the first scene, the buildings in the red crop have low contrast in both Retinex-MEF and Diff-MEF. Although building contours are clearer after AnyMatch pre-registration, both fused images show a dark band at the right edge. GenRF renders clearer facades with more uniform exposure across the skyline and a seamless transition into the non-overlapping region at the edge. In the second scene, both Retinex-MEF and Diff-MEF yield low-contrast results even after pre-registration. It is worth noting that directly fusing the misaligned inputs produces visible ghosting around the door and air-conditioning unit in the green crop. GenRF improves local contrast directly from the misaligned inputs, rendering clear contours around the door and air-conditioning unit without visible ghosting. GenRF also retains the poles and diagonal supports visible in the red crop. In the third scene, pre-registration reduces the duplicated building edges in both alternatives, yet the foliage in Retinex-MEF as well as the windows and wall in the green crop of Diff-MEF still have low contrast. GenRF directly fuses the misaligned pair and clearly delineates both the window frames and foliage. The above examples demonstrate that GenRF leverages the pretrained FLUX prior to achieve balanced exposure, stronger local contrast, and natural colors.

\textbf{Multi-focus image fusion.} We use the test instruction: \inlineprompt{Fuse the two differently focused images of the same scene into a sharp all-in-focus RGB image. Preserve scene geometry, realistic colors, and fine details in both near and far focus regions. Avoid ghosting, halos, defocus blur, and invented details}, for multi-focus image fusion. As shown in Fig.~\ref{fig:supp-mff}, misalignment produces duplicated contours in UHD-MFF and causes ND-CNPFuse to misidentify in-focus regions, leaving holes in its fused images. In the first scene, AnyMatch reduces artifacts along the red frames and building edges, but shadow details in the red crop remain hard to discern. GenRF combines details from different focus planes into an all-in-focus image and brightens the shadowed region. In the second scene, the foreground object in the green crop remains blurred even in the near-focus input. This residual blur persists in the pre-registered results of both competing methods. The pretrained FLUX prior helps GenRF render the foreground details more clearly. In the third scene, visible artifacts remain around the figurine in the green crop after pre-registration. GenRF combines in-focus details from both source images, keeping the figurine and the background window sharp. It also brightens the area around the window, making the bars more visible in the red crop. The aforementioned examples demonstrate that GenRF can combine in-focus details across misaligned inputs while reducing residual blur and improving exposure.

\begin{figure}[t]
    \centering
    \includegraphics[width=\linewidth]{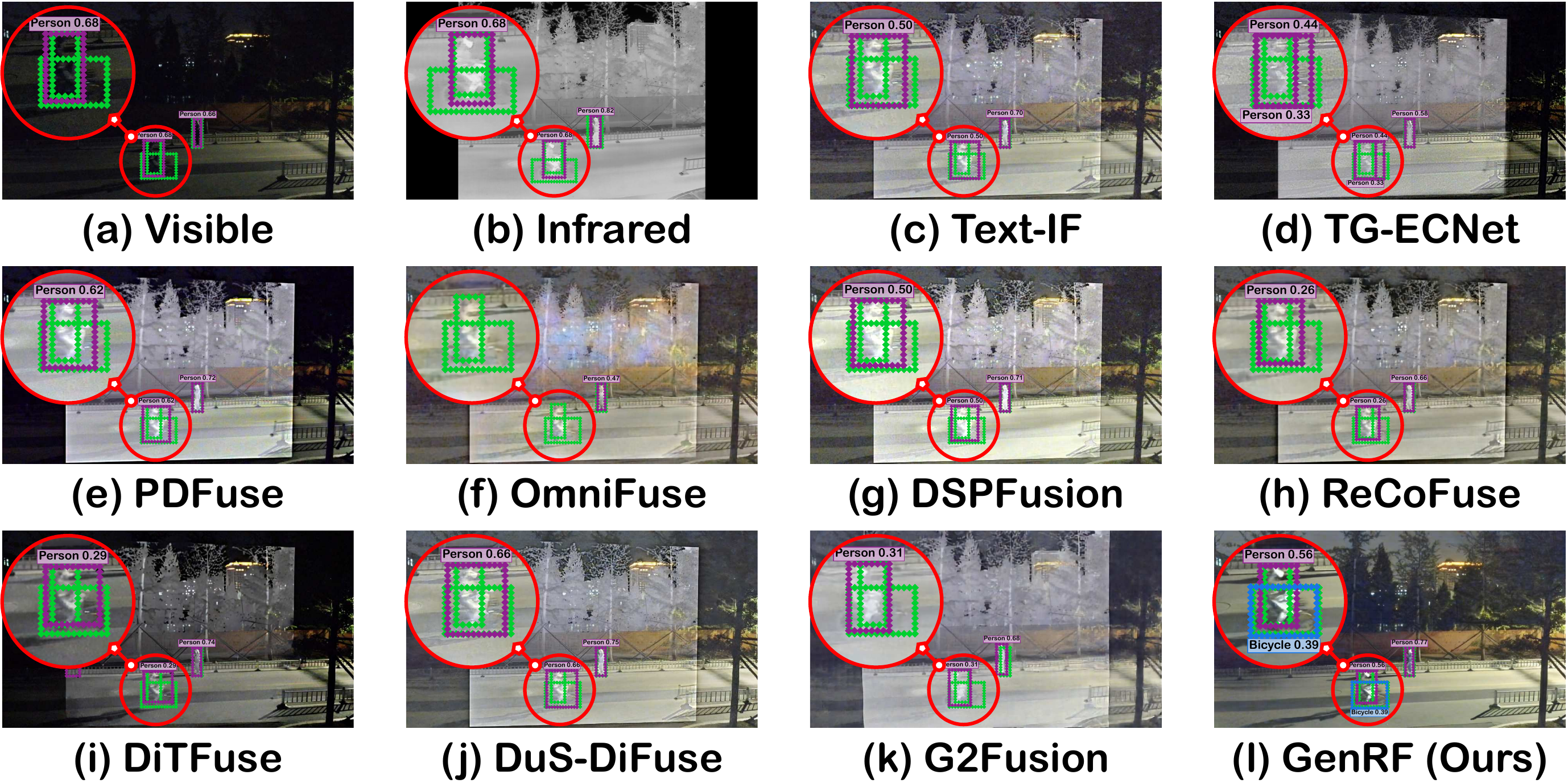}
    \caption{Qualitative object detection comparison on DVTOD. Dashed boxes \detbox{detgreen}, \detbox{detpurple}, and \detbox{detblue} denote ground truth, person predictions, and bicycle predictions, respectively.}
    \label{fig:supp-detection}
    \vspace{-0.1in}
\end{figure}

\section{Qualitative Object Detection Results}
Figure~\ref{fig:supp-detection} presents predictions from the same pretrained YOLO-11s detector on the source images and fusion results. Green dashed boxes indicate ground-truth annotations in visible coordinates. Purple and blue dashed boxes indicate person and bicycle predictions, respectively. As shown in the figure, the detector identifies the person in most fused images but misses the adjacent bicycle. OmniFuse misses both objects, and TG-ECNet produces an additional person detection near the bicycle. GenRF preserves the pedestrian’s outline and makes the adjacent bicycle easier to discern, allowing the detector to identify both objects.

\section{Discussion on Computational Cost}
\label{sec:supp-inference-cost}

Although GenRF achieves promising fusion results on real captures, its computational cost remains a major challenge. Cross-attention between source images increases the cost per denoising step, while leveraging the pretrained FLUX generative prior requires multiple denoising steps. As shown in Table~\ref{tab:supp-msrs-cost}, diffusion-based methods such as PDFuse, DiTFuse, DuS-DiFuse, and GenRF have comparatively large parameter counts. In particular, GenRF takes 23.981~s to fuse a $640\times480$ image pair over 30 denoising steps, which may limit its deployment in real-time systems. Therefore, future work could explore one-step generative fusion with a feature distribution matching objective~\cite{supp-Zhang2026MGFlow} to reduce inference latency.

\begin{table}[t]
\centering
\small
\setlength{\tabcolsep}{4pt}
\caption{Computational cost comparison for $640\times480$ inputs.
Rows marked with
\protect\colorbox[HTML]{F2F2F2}{\protect\rule{0pt}{0.55ex}\protect\hspace{1em}}
report the cost of joint restoration and fusion.
The full pipeline cost for each of these methods also includes the separately reported AnyMatch pre-registration cost.}
\label{tab:supp-msrs-cost}
\begin{tabularx}{\columnwidth}{>{\raggedright\arraybackslash}Xrrr}
\toprule
\textbf{Method} & \textbf{Params (M)} & \textbf{GFLOPs} & \textbf{Time (s)} \\
\midrule
\rowcolor[HTML]{D9D9D9}
AnyMatch & 415.66 & 6,576.96 & 0.487 \\
\midrule
\rowcolor[HTML]{F2F2F2}
Text-IF & 215.12 & 3,240.14 & 0.250 \\
\rowcolor[HTML]{F2F2F2}
TG-ECNet & 40.22 & 6,980.73 & 0.742 \\
\rowcolor[HTML]{F2F2F2}
PDFuse & 1,066.24 & 538,283.39 & 11.200 \\
\rowcolor[HTML]{F2F2F2}
OmniFuse & 173.34 & 3,219.58 & 5.607 \\
\rowcolor[HTML]{F2F2F2}
DSPFusion & 14.01 & 564.26 & 0.253 \\
\rowcolor[HTML]{F2F2F2}
ReCoFuse & 155.47 & 3,786.15 & 5.978 \\
\rowcolor[HTML]{F2F2F2}
DiTFuse & 3,845.73 & 1,557,315.46 & 26.069 \\
\rowcolor[HTML]{F2F2F2}
DuS-DiFuse & 3,391.62 & 106,078.71 & 4.204 \\
\midrule
\rowcolor[HTML]{EAF5EA}
G2Fusion & 778.51 & 13,769.70 & 2.063 \\
\rowcolor[HTML]{EAF5EA}
GenRF (Ours) & 5,662.66 & 2,429,833.52 & 23.981 \\
\bottomrule
\end{tabularx}
\end{table}

% \clearpage

\end{document}